\documentclass[letterpaper, 10 pt, conference]{ieeeconf}  

\IEEEoverridecommandlockouts                              

\usepackage[noadjust]{cite}
\usepackage{graphicx}
\usepackage{amsmath}
\usepackage{amssymb}
\usepackage{booktabs}
\usepackage{multirow}
\usepackage{caption}
\usepackage{subcaption}

\usepackage[capitalize]{cleveref}
\crefname{section}{Sec.}{Secs.}
\Crefname{section}{Section}{Sections}
\Crefname{table}{Table}{Tables}
\crefname{table}{Tab.}{Tabs.}

\title{\LARGE \bf
Video Killed the HD-Map: \\ Predicting Multi-Agent Behavior Directly From Aerial Images
}

\author{
Yunpeng Liu$^{1,2}$
Vasileios Lioutas$^{1,2}$
Jonathan Wilder Lavington$^{1,2}$
Matthew Niedoba$^{1,2}$
Justice Sefas$^{1,2}$\\
Setareh Dabiri$^{1}$
Dylan Green$^{1,2}$
Xiaoxuan Liang$^{1,2}$
Berend Zwartsenberg$^{1}$ 
Adam \'Scibior$^{1}$
Frank Wood$^{1,2,3}$
\thanks{$^{1}$Inverted AI,
$^{2}$University of British Columbia,
$^{3}$Mila}
}

\begin{document}

\maketitle
\thispagestyle{empty}
\pagestyle{empty}

\begin{abstract}
The development of algorithms that learn multi-agent behavioral models using human demonstrations has led to increasingly realistic simulations in the field of autonomous driving. In general, such models learn to jointly predict trajectories for all controlled agents by exploiting road context information such as drivable lanes obtained from manually annotated high-definition (HD) maps. Recent studies show that these models can greatly benefit from increasing the amount of human data available for training. However, the manual annotation of HD maps which is necessary for every new location puts a bottleneck on efficiently scaling up human traffic datasets. We propose an aerial image-based map (AIM) representation that requires minimal annotation and provides rich road context information for traffic agents like pedestrians and vehicles. We evaluate multi-agent trajectory prediction using the AIM by incorporating it into a differentiable driving simulator as an image-texture-based differentiable rendering module. Our results demonstrate competitive multi-agent trajectory prediction performance especially for pedestrians in the scene when using our AIM representation as compared to models trained with rasterized HD maps. 
\end{abstract}

\section{Introduction}
\label{sec:intro}
Creating realistic simulation environments is crucial for evaluating self-driving vehicles before they can be deployed in the real world. Recent studies have emphasized the use of learned models to generate more realistic  behavior for controlled agents like pedestrians and surrounding vehicles~\cite{liu2022vehicle,suo2021trafficsim,9565113}. These models learn to imitate human-like behaviors in a traffic scene by utilizing a probabilistic conditional model of multi-agent trajectories in an environment. When using such approaches to construct realistic simulations, the quality of learned behavior is strongly dependent on the amount of data used for training~\cite{ettinger2021large, houston2021one}.
\begin{figure}[tp]
     \centering
     \begin{subfigure}[b]{0.45\columnwidth}
         \centering
         \includegraphics[width=\columnwidth]{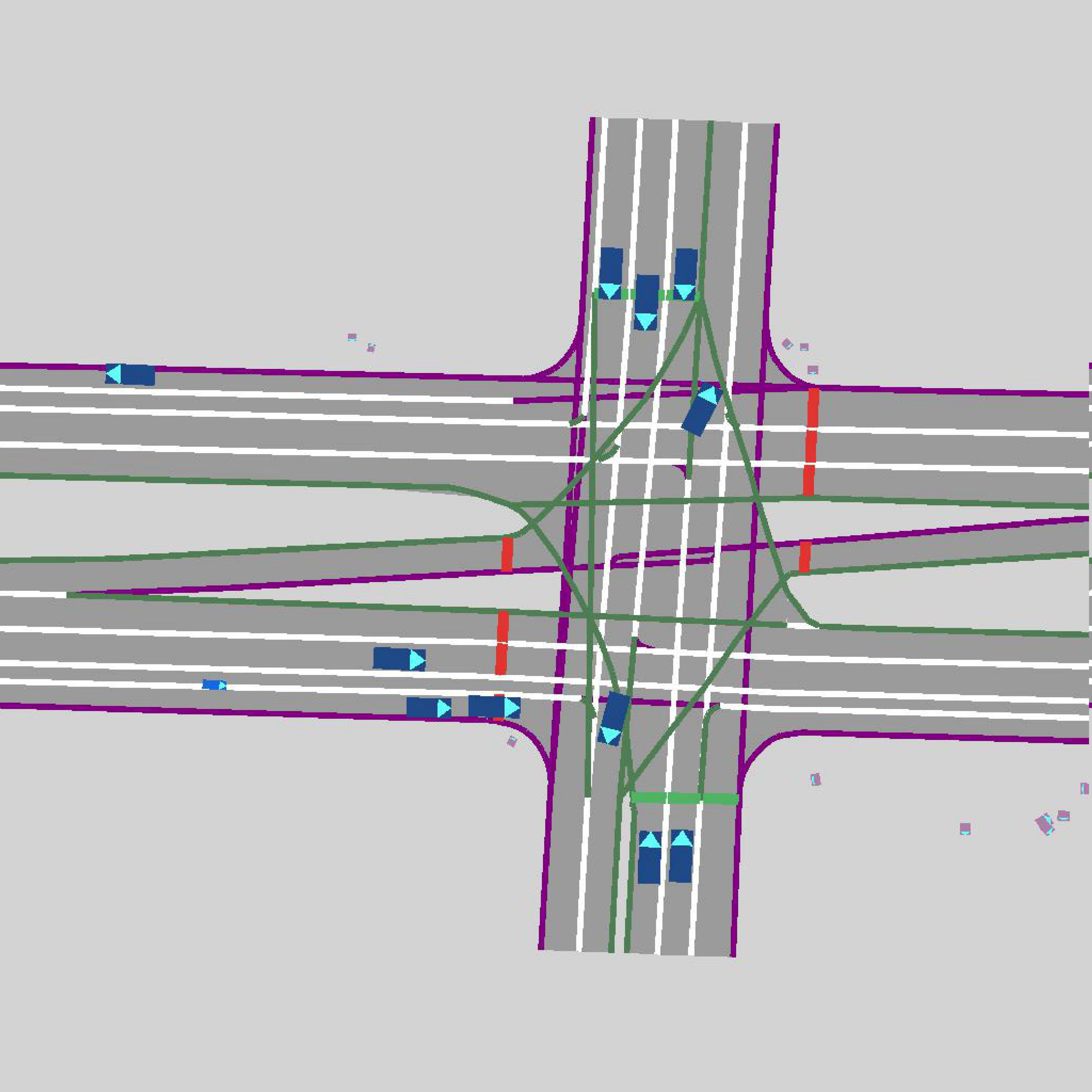}
         \caption{}
         \label{fig:birdview}
     \end{subfigure}
     \hfill
     \begin{subfigure}[b]{0.45\columnwidth}
         \centering
         \includegraphics[width=\columnwidth]{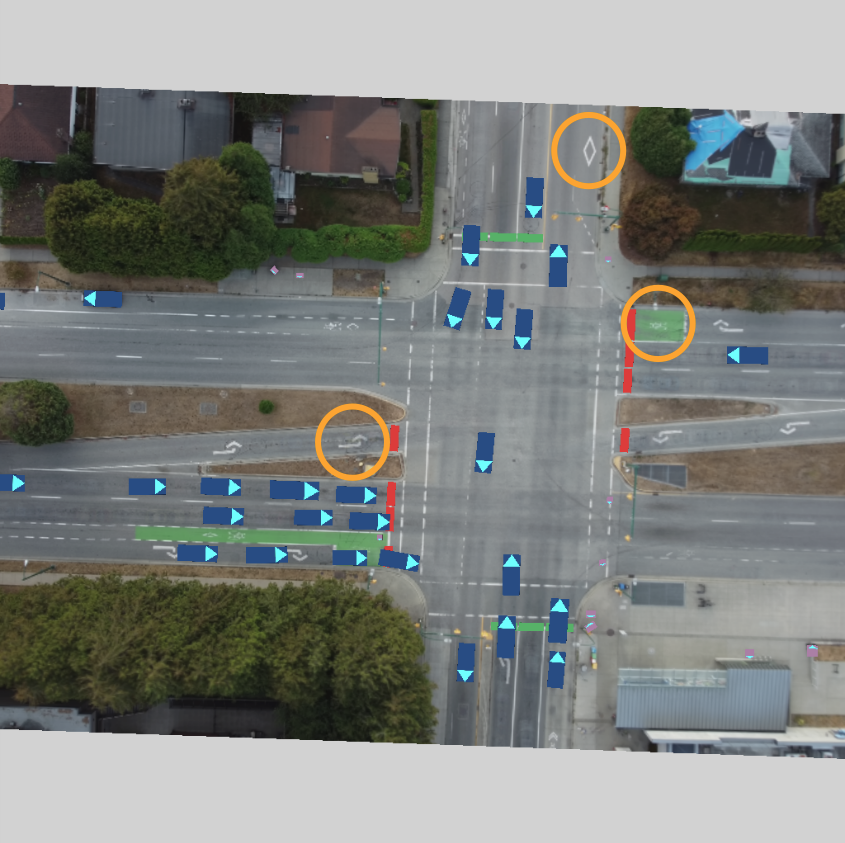}
         \caption{}
         \label{fig:text_birdview}
     \end{subfigure} 
          \hfill
    \begin{subfigure}[b]{0.45\columnwidth}
         \centering
         \includegraphics[width=\columnwidth]{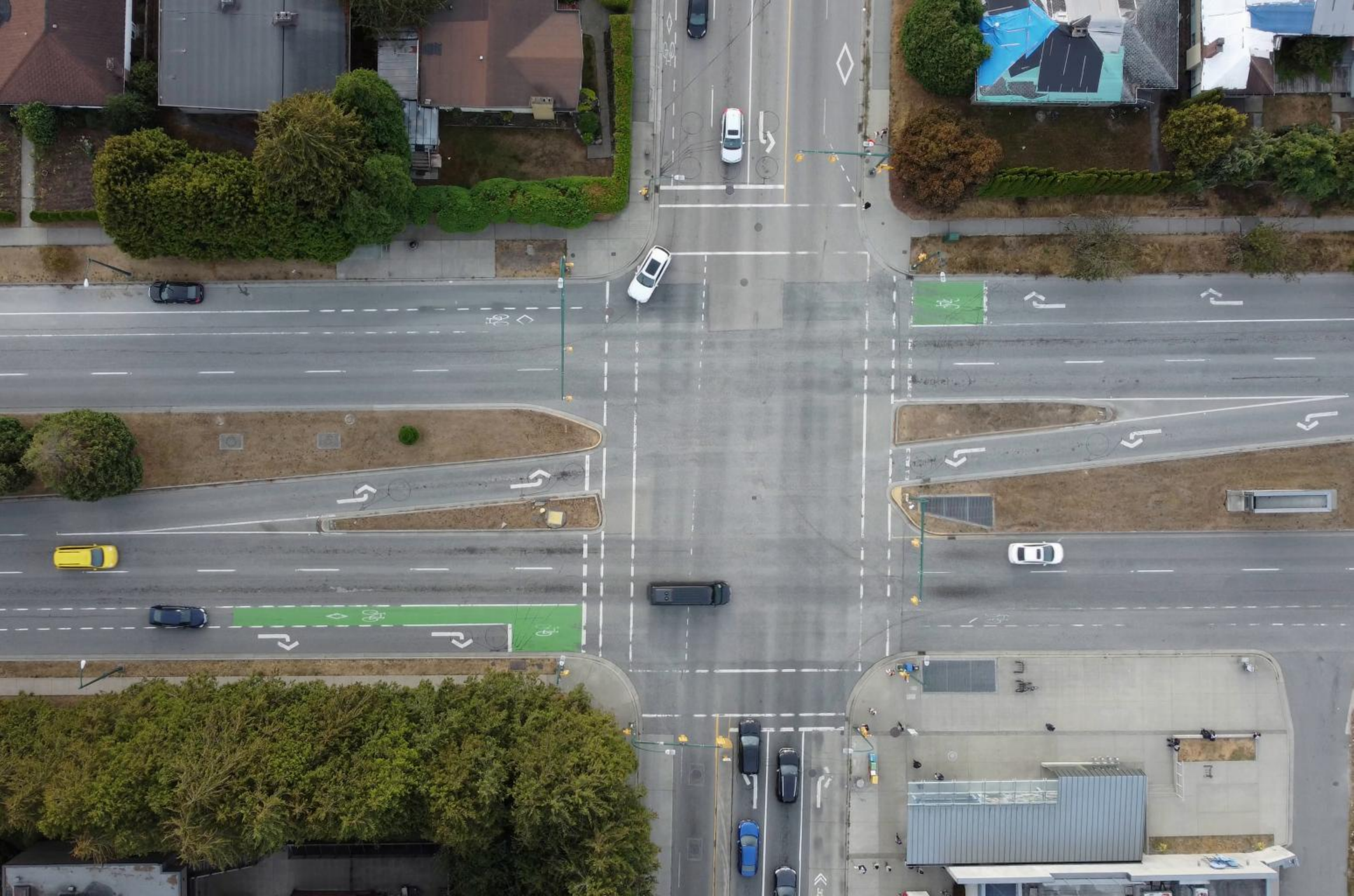}
         \caption{}
         \label{fig:backimg_car}
     \end{subfigure} 
     \hfill
        \begin{subfigure}[b]{0.45\columnwidth}
         \centering
         \includegraphics[width=\columnwidth]{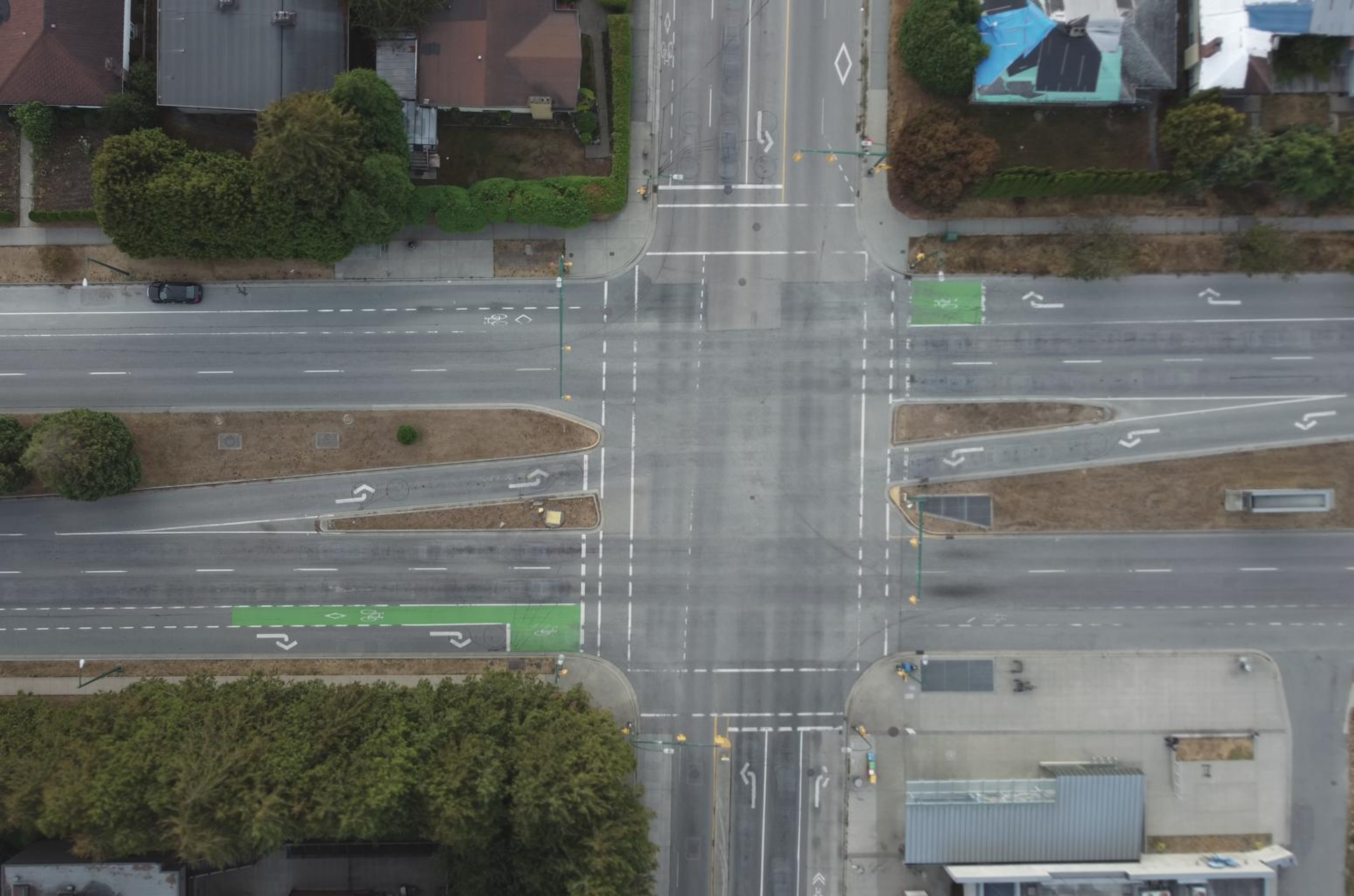}
         \caption{}
         \label{fig:backimg}
     \end{subfigure}
     \caption{(a) An example of a simulated scene with the rasterized HD map representation compared to (b) the aerial image-based map (AIM) representation rendered using our image-texture-based differentiable rendering module. The AIM representation requires minimum annotation effort as it is obtained directly from (c) the raw drone video recording frame with agents removed (d). Orange circles in (b) highlight examples of rich road context information.}
\end{figure}

\begin{figure}[t]
     \centering
     \begin{subfigure}[b]{0.45\columnwidth}
         \centering
         \includegraphics[width=\columnwidth]{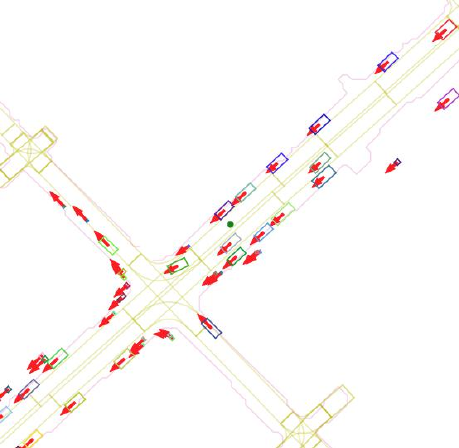}
         \caption{}
         \label{fig:argoverse}
     \end{subfigure}
     \hfill
     \begin{subfigure}[b]{0.45\columnwidth}
         \centering
         \includegraphics[width=\columnwidth]{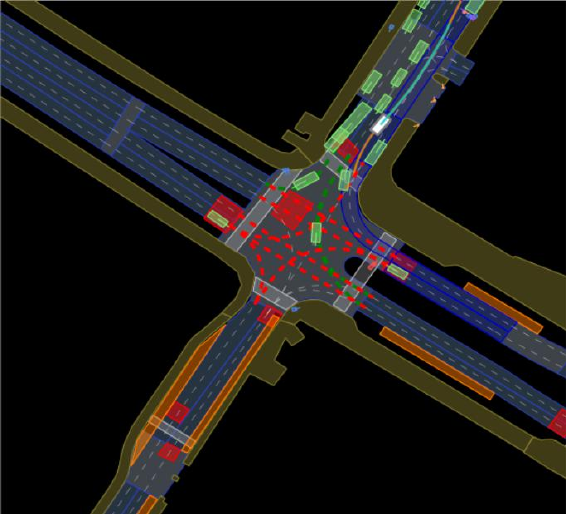}
         \caption{}
         \label{fig:nuplan}
     \end{subfigure}
     \caption{Examples of HD maps from public motion planning datasets for (a) Argoverse~\cite{Argoverse} and (b) Nuplan. The Nuplan map includes annotations like crosswalks and parking areas (shown in orange), which are not labeled in the Argoverse map.}
     \label{fig:maps}
\end{figure}

Typically, learning behavior models requires data consisting of the high-definition (HD) map for the given location and extracted agent tracks.  While the latter can be extracted automatically from sensory data using modern computer vision algorithms with good accuracy~\cite{caesar2021nuplan,ettinger2021large,houston2021one}, doing that for the former is still an open problem~\cite{elhousni2020automatic,Li2021HDMapNetAO,9636205} and in practice, manual annotations are often used. Moreover, since it is important to ensure not only a large number of hours in the dataset but also a large variety of locations, manually annotating HD maps can become the most laborious part of creating a dataset. To make things worse, HD maps inevitably fail to capture important context, and increasing their detail like annotating sidewalks and crosswalks (see \cref{fig:nuplan}) increases the cost of annotation. For example, the simplistic HD map scheme used in \cref{fig:birdview} does not reflect pedestrian crosswalks, sidewalks and bus lane designations. 

In this study, we investigate the performance of behavioral models learned using aerial imagery instead of HD maps. Specifically, we record a dataset of human behavior in traffic scenes with a drone from a bird's-eye view, in a manner similar to~\cite{interactiondataset}, and extract the background aerial image by averaging the collected video frames of the location. While other background extraction techniques can be applied~\cite{li2018adaptive,zhang2012nonparametric,barnich2010vibe}, we find this simple averaging approach is sufficient for our use case. We refer to this image as the ``aerial image-based map'' (AIM), emphasizing that it is both easy to obtain automatically and that it contains rich contextual information. 


Learning trajectory prediction models by behavioral cloning is known to suffer from the covariate shift, where prediction quality drops drastically with simulation time, and it has been demonstrated that this issue can be ameliorated by imitating in a differentiable simulator instead~\cite{suo2021trafficsim,9565113,scheel2022urban}. We use a similar approach incorporating the AIM into a differentiable simulator by implementing a custom differentiable renderer. The renderer, illustrated in \cref{fig:rendering_procedure}, uses the AIM as background and places simple rasterizations of agents and traffic lights on top of it, as shown in \cref{fig:text_birdview}. To evaluate the impact of using AIM, we employ a multi-agent trajectory prediction model, ITRA~\cite{9565113} which consumes rasterized views of HD maps shown in \cref{fig:birdview}. We compare ITRA trained with AIM representation (ITRA-AIM) shown in \cref{fig:text_birdview} with the same model using HD map representation (ITRA-HDM) for two dominant traffic agent categories, pedestrians and vehicles. ITRA-AIM demonstrates competitive performance compared to ITRA-HDM on widely used metrics such as minimum Average Displacement Error (minADE) and minimum Final Displacement Error (minFDE) for both agent types. Notably, ITRA-AIM exhibits an even higher performance gain on pedestrian trajectory prediction.


\section{Background}
\label{sec:background}
In this section, we will formally define our multi-agent trajectory prediction problem and give an overview of ITRA, the multi-agent trajectory prediction model which we use to evaluate our AIM representation. We will also introduce  the concept of differentiable driving simulators which are applied in many trajectory prediction models, including ITRA.

\subsection{Multi-agent trajectory prediction}
\label{sec:MATP}
In this paper, we define the state for $N$ agents across $T$ time steps as $s^{N}_{T}$ (following the notation used in~\cite{9565113}). For a specific agent $i$, its state $s^{i}_{t}= (x^i_t, y^i_t, \phi^i_t, v^i_t)$ for $t \in 1,\dots,T$ consists of the agent's coordinates, as well as its direction and velocity relative to a stationary global reference frame. In the multi-agent trajectory prediction setting, we are interested in predicting the future joint state $s_{t_{obs+1}:T}^{1:N}$, given $t_{obs}$ state observations while conditioning on the road context information. This information is traditionally represented by a so-called HD map. These maps consist of road polygons, lane boundaries, lane directions, and may also include additional features such as crosswalks.


\subsection{ITRA}
We use ITRA~\cite{9565113} to investigate the validity of our primary claim. ITRA uses a conditional variational recurrent neural network (CVRNN)~\cite{chung2015recurrent} model followed by a bicycle kinematic model~\cite{7995816} to jointly predict the next state of each agent in the scene. All interactions between the agents and the environment are encoded using differentiably rendered birdview images. These birdview representations are centered at the agent of interest and rotated to match its orientation. Each agent $i$ at timestep $t$ is modeled as a CVRNN with recurrent state $h_t^i$ and latent variables $z_t^i \sim \mathcal{N}(z_t^i; 0, \mathbf{I})$. After ITRA obtains the corresponding ego-centered birdview $b_t^i = \mathtt{render}(i, s_t^{1:N_t})$, it produces the next action $a_t^i = f_\theta(b_t^i, z_t^i, h_{t-1}^i)$, where $h_t^i = f_{\psi}(h_{t-1}^i, b_t^i, a_t^i)$ is generated using a recurrent neural network. The next state $s_{t+1}^i \sim \mathcal{N}(s_{t+1}^i;f_{kin}(s^i_t, a^i_t), \sigma\mathbf{I})$ is produced using a kinematic bicycle model $f_{kin}$ and the generated action $a_t^i$. The joint model $p(s_{1:T}^{1:N})$ factorizes as
\begin{align}
\int\!\!\int\!\prod^{T}_{t=1}\!\prod^{N}_{i=1}\!  p(s^i_{t+1}|s^i_{t}, a^{i}_{t})p_{\theta}(a_t^i| b_t^i, &z_t^i, h_{t-1}^i) \nonumber \\[-10pt]&p(z^i_t) dz_{1:T}^{1:N}da_{1:T}^{1:N}.   
\end{align}
The model is trained jointly with a separate inference network $q_\phi(z_t^i | b_t^i, a_t^i, h_{t-1}^i)$ by maximizing the evidence lower bound (ELBO),
\begin{align}
\sum^{T}_{t=1}\sum^{N}_{i=1} &\mathbb{E}_{q_\phi(z_t^i | b_t^i, a_t^i, h_{t-1}^i)}\left[\log p_{\theta}(s^i_{t+1}|b_t^i, z_t^i, h_{t-1}^i)\right] \nonumber\\
-& \mathrm{KL}\left[ q_\phi(z_t^i | b_t^i, a_t^i, h_{t-1}^i)|| p(z^i_t)\right],
\label{eq:elbo}
\end{align} where 
\vspace{-1em}
\begin{align}
 p_{\theta}(s^i_{t+1}|b_t^i, z_t^i, h_{t-1}^i) \!=\!\! \int\! p(s^i_{t+1}|s^i_{t}, a^{i}_{t})p_{\theta}(a_t^i| b_t^i, z_t^i, h_{t-1}^i) d_{a^i_t}. \nonumber
\end{align}    

\subsection{Differentiable driving simulators}
Previous research has demonstrated that performing imitation learning within a differentiable simulator can help mitigate the distributional shift due to compounded error in open-loop behavior cloning methods \cite{ross2011reduction,scheel2022urban,suo2021trafficsim}. These simulators typically consist of a differentiable kinematic model,  which produces the next state $s^i_{t+1}$ given the current state-action pair. Additionally, they have a differentiable renderer that generates the ego-centered bird's-eye view image that includes the road context information and other agents in the scene. One of the main advantages of using such differentiable simulators is that the loss in \cref{eq:elbo} can be directly optimized using backpropagation as the state transition $p(s^i_{t+1}|s^i_{t}, a^{i}_{t})$ is fully differentiable.





\section{Related Work}
\label{sec:related}
Current methods to multi-agent modeling approach the problem by jointly predicting future trajectories using deep probabilistic models such as conditional variational auto-encoders (CVAEs)~\cite{Lee2017DESIREDF, Tang2019MultipleFP, 9565113, suo2021trafficsim}, normalizing flows~\cite{Rhinehart_2019_ICCV,ma2020normalizing} and more recently diffusion models~\cite{zhong2022guided}. This family of multi-agent trajectory prediction models relies heavily on obtaining road context from HD maps, which requires manual annotations of lane center and boundary lines. To represent such HD maps, one approach is to render the semantic information of the map into a birdview image by employing different color schemes ~\cite{9565113,suo2021trafficsim,scheel2022urban}. This image is then encoded using a convolutional neural network (CNN).  Alternatively, recent work~\cite{liang2020learning,Gao_2020_CVPR} suggests representing road elements as a sequence of vectors that can be employed by a graph neural network (GNN) which achieves better performance than the rendering approach. However, regardless of the embedding method, an annotated map has to be obtained, which our method bypasses completely. Moreover, aerial images contain rich road context information such as left and right turn road markings and bus lanes without any annotation effort. In prior research~\cite{zhang2022trajectory}, satellite image-based maps were used as a substitute for HD semantic maps, and the resulting findings indicated that trajectory prediction exhibited worse performance in such images. We point out that in those cases, satellite map representation that contains traffic light states were not evaluated and the visual quality of the road context is limited. In contrast, our AIM representation does contain traffic light states, and the representation of agent states obtained from our differentiable rendering module has a higher visual quality. We demonstrate in \cref{sec:ablation} that these two factors can have significant impact on the prediction performance.

Previous studies have shown that integrating scene context into pedestrian trajectory prediction models improves their performance~\cite{korbmacher2022review}.  These studies~\cite{sadeghian2019sophie,zhao2019multi} often encode a single static bird's-eye view image of the scene using a CNN to represent scene information. The sequence of video frames can also serve as input for scene context information using graph convolutional neural networks~\cite{mohamed2020social}. 
To incorporate agent information, these methods utilize a separate network, and scene information is merged with agent information using attention~\cite{sadeghian2019sophie}, GNNs~\cite{mohamed2020social}  or CNNs~\cite{zhao2019multi}. However, the aforementioned methods suffer from covariate shift over a long time, limiting their application for simulation purposes.

\section{Proposed Method}
\label{sec:method}
Our method incorporates unlabelled aerial images into a simulation environment~\cite{9565113} using a differentiable renderer implemented with Pytorch3D~\cite{ravi2020pytorch3d}. We leverage image-texture-based rendering to represent the background of the simulated scene, as depicted in \cref{fig:text_birdview}. By embedding this rendering module into an existing end-to-end differentiable 2D driving simulator that targets multi-agent trajectory prediction, we can produce ego-rotated and egocentric birdview images $b_t^i$ that utilize the proposed aerial image-based map (AIM) for multi-agent trajectory prediction models like ITRA. Furthermore, it provides a representation of the road context with minimal information loss, as opposed to a rasterized birdview image from the labeled HD map shown in \cref{fig:birdview}.  We introduce our image-texture-based differentiable rendering module in the following section.

\begin{figure}[t]
     \centering
     \begin{subfigure}[b]{\columnwidth}
         \centering
         \includegraphics[width=\columnwidth]{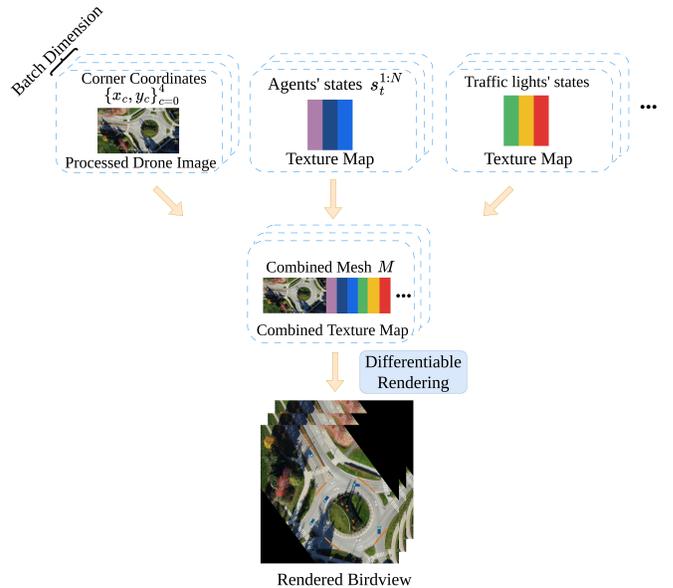}
     \end{subfigure}
    \caption{Image-texture-based rendering procedure.}
    \label{fig:rendering_procedure}
\end{figure}
\subsection{Image-texture-based differentiable rendering.}
Our image-texture-based rendering module is designed to be differentiable and efficient as it supports rendering in batch mode. The rendering process is illustrated in \cref{fig:rendering_procedure}. Our module takes in the processed drone image of the recording location, along with its coordinates of the four image corners ${\{x_c, y_c\}}^{4}_{c=0}$,
the states of the agents $s_{t}^{1:N}$ at time $t$, and the traffic light states. We average motion stabilized drone video frames to perform a simple yet effective background extraction of the video. This background extraction process serves to eliminate cars and other agents that may be present in the drone video.  To differentiate agent types, each agent type is associated with a unique color in the texture map (see \cref{fig:rendering_procedure}). The image corner coordinates are in the same global reference frame as the agent coordinates to align the map with the agents during the rendering process. Using the aforementioned inputs, three distinct types of meshes are constructed, namely the background mesh, agent mesh and the traffic light mesh along with their corresponding texture maps. These meshes are subsequently combined to form a concatenated mesh $M$ with a merged texture map,  which is then fed into a differentiable renderer, to render the simulated scene.  Agents are rendered as bounding boxes with an additional triangle on top of each bounding box to indicate their direction. Traffic lights are rendered as rectangular bars at the stop line. The rendered birdview can be consumed by the trajectory prediction models as a representation of the environment for the ego agent which provides information about the road context and other agents in the scene. 

\section{Experiments}
\label{sec:eval}
\begin{table}[t]
    \centering
        \caption{Validation set prediction errors on our pedestrian dataset.}
    \begin{tabular}{l||ccc}
    \hline 
    Method & $\text{minADE}_6$$\downarrow$ & $\text{minFDE}_6$$\downarrow$ & $\text{MFD}_6$$\uparrow$\\ 
    \hline
    ITRA-P-HDM & 0.90 & 2.01 & 0.29\\
    ITRA-P-AIM & \bf{0.74} & \bf{1.56}& \bf{0.51}\\
    \hline
    \end{tabular}
    \label{tab:p_valid}
\end{table}

    

\begin{table}[t]
    \centering
    \caption{Validation set prediction errors on our vehicle dataset.}
    \resizebox{\columnwidth}{!}{%
    \begin{tabular}{l||ccc}
        \hline
        Metrics & ITRA-V-HDM & ITRA-V-AIM & ITRA-V-AIM-ResNet18 \\
        \hline
        $\text{minADE}_6$$\downarrow$ &0.50 & \bf{0.45}&  0.50\\
        $\text{minFDE}_6$$\downarrow$ & 1.04 &\bf{0.93} & 1.10 \\
        Off-road rate$\downarrow$ &\bf{0.006} &  0.008& \bf{0.006} \\
                collision rate$\downarrow$ & \bf{0.012}&\bf{0.012} &\bf{0.012} \\
        \hline
    \end{tabular}%
    }
    \label{tab:v_valid}
\end{table}


\begin{figure*}[t]
     \centering
     \begin{subfigure}[b]{0.18\textwidth}
        \centering
        \includegraphics[width=1\linewidth,keepaspectratio]{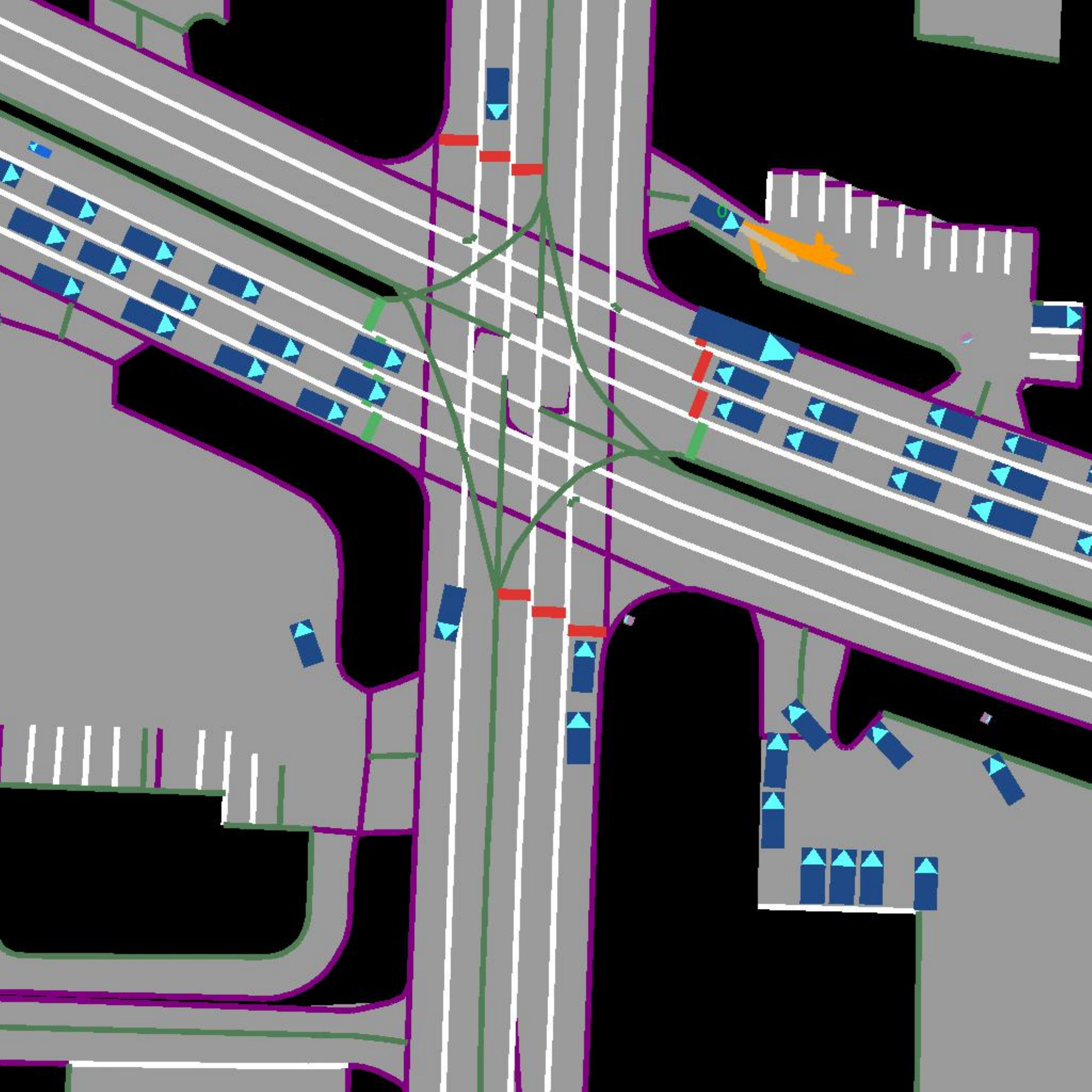}
    \end{subfigure}
    \begin{subfigure}[b]{0.18\textwidth}
        \centering
        \includegraphics[width=1\linewidth,keepaspectratio]{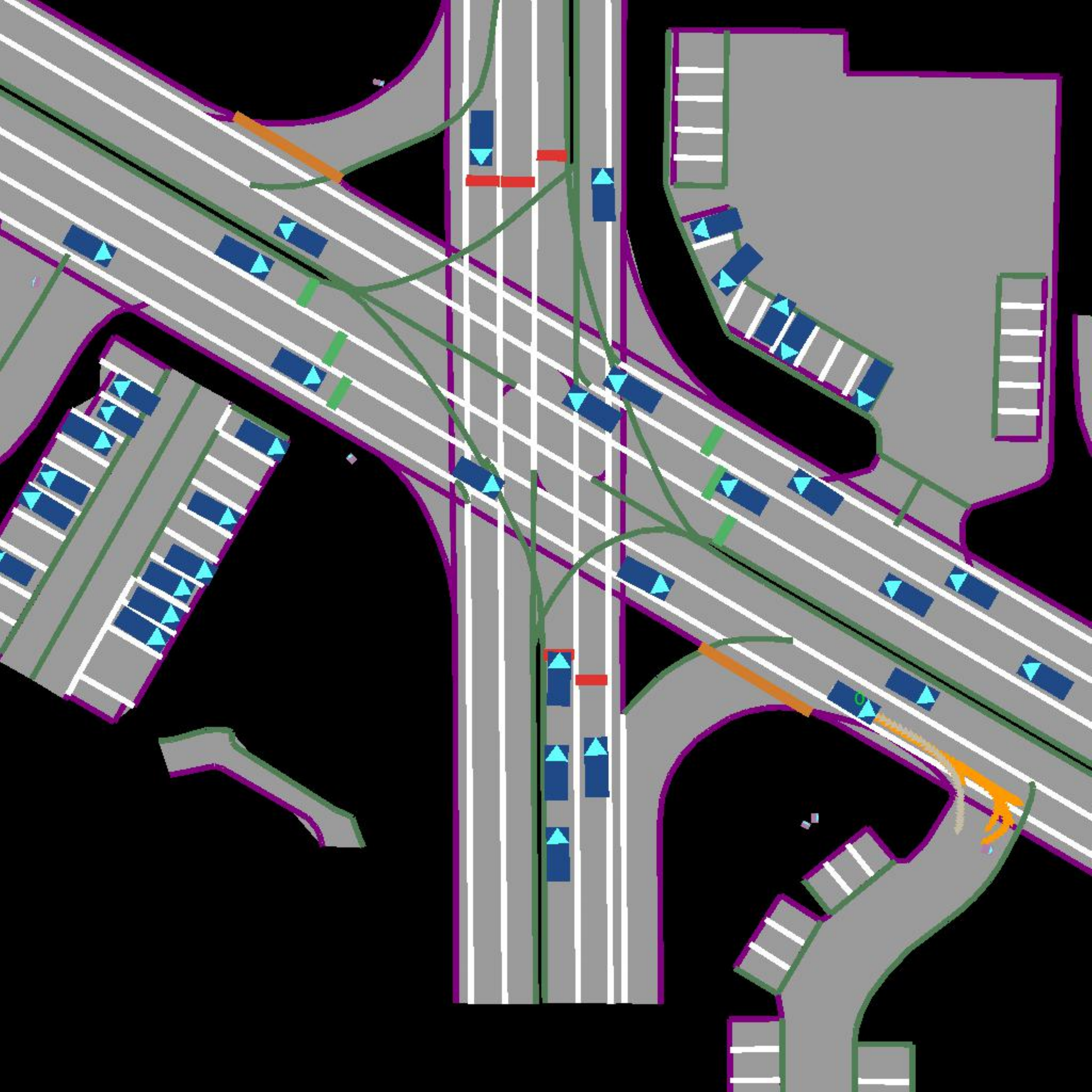}
    \end{subfigure}
    \begin{subfigure}[b]{0.18\textwidth}
        \centering
        \includegraphics[width=1\linewidth,keepaspectratio]{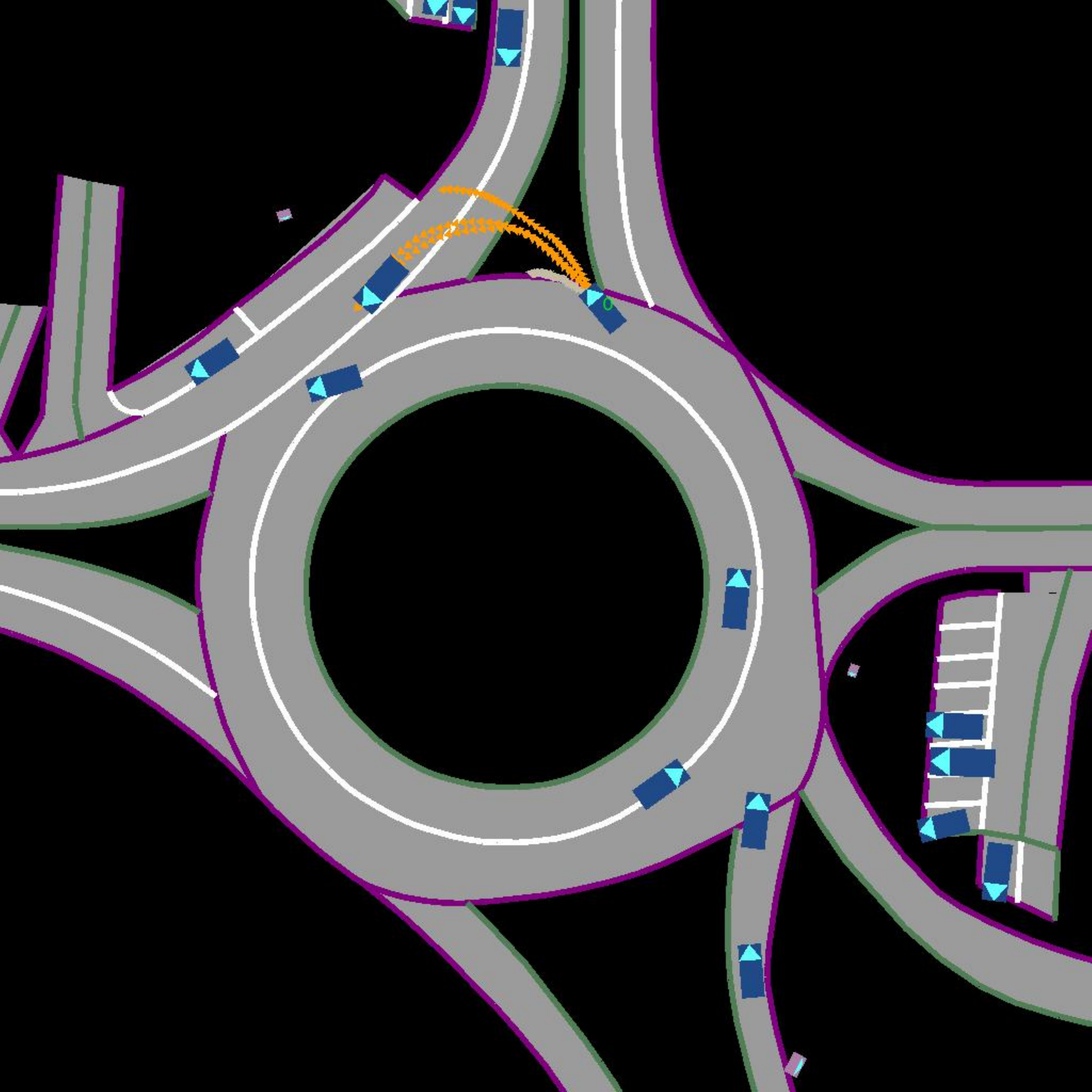}
    \end{subfigure}
        \begin{subfigure}[b]{0.18\textwidth}
        \centering
        \includegraphics[width=1\linewidth,keepaspectratio]{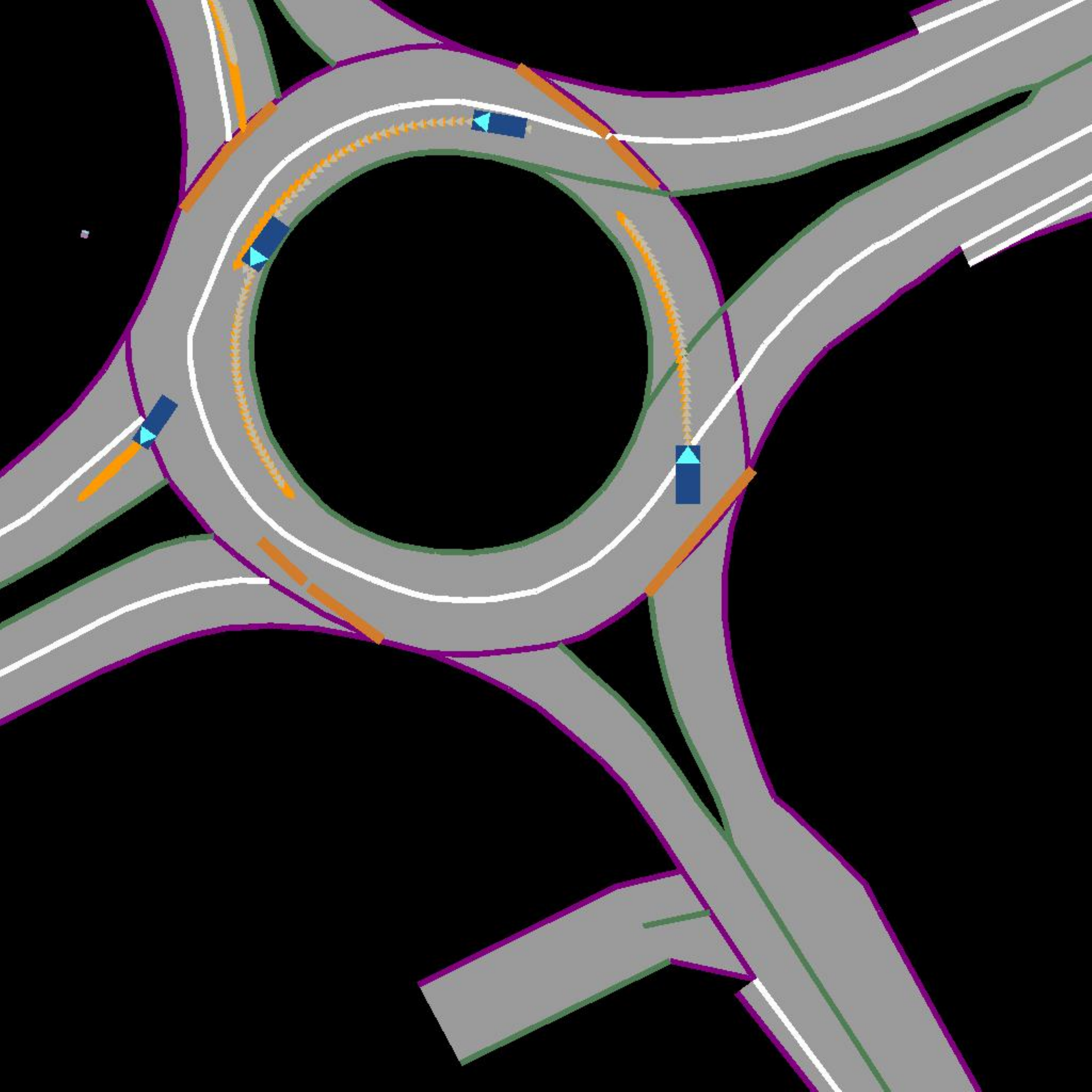}
    \end{subfigure}
        \begin{subfigure}[b]{0.18\textwidth}
        \centering
        \includegraphics[width=1\linewidth,keepaspectratio]{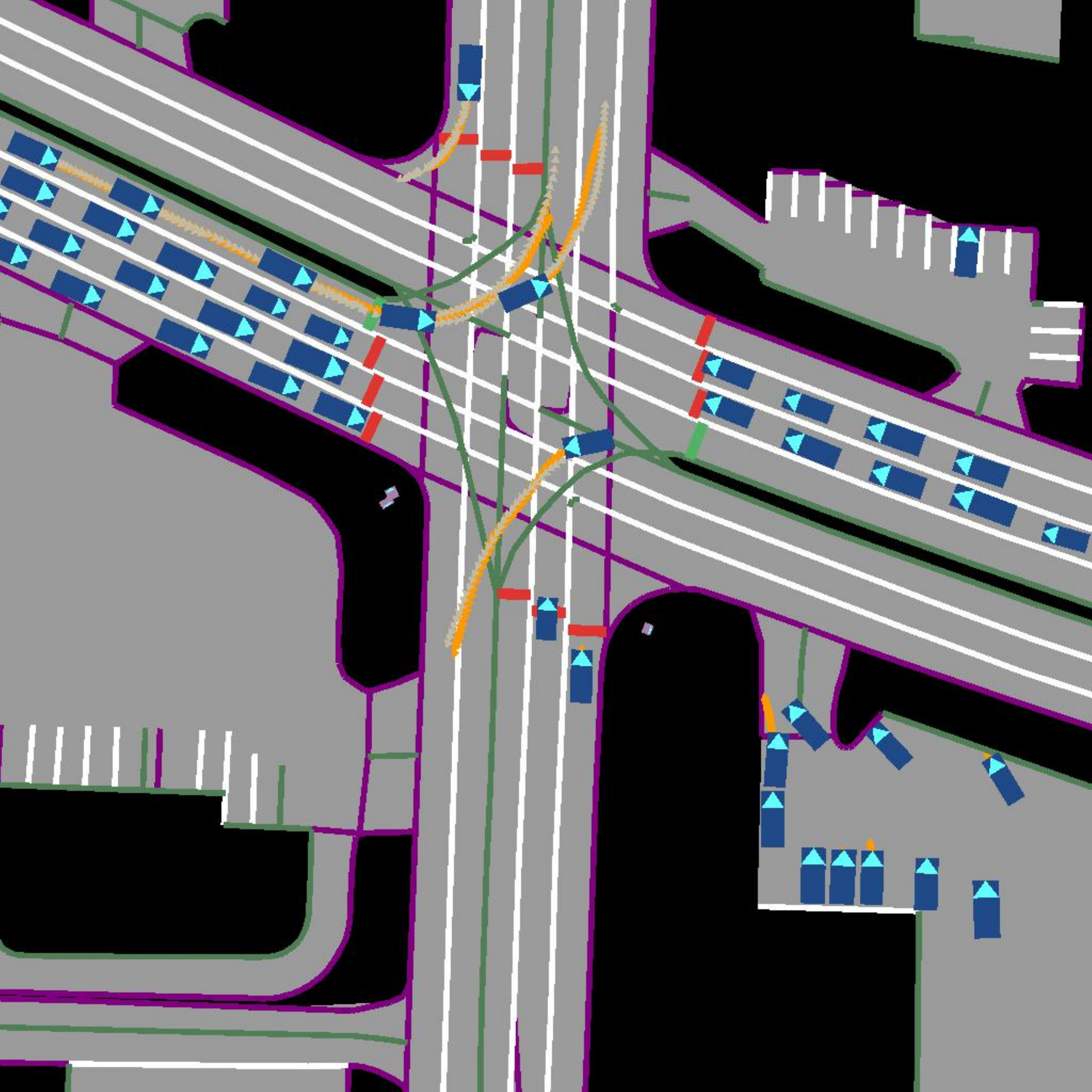}
    \end{subfigure}
            \begin{subfigure}[b]{0.18\textwidth}
        \centering
        \includegraphics[width=1\linewidth,keepaspectratio]{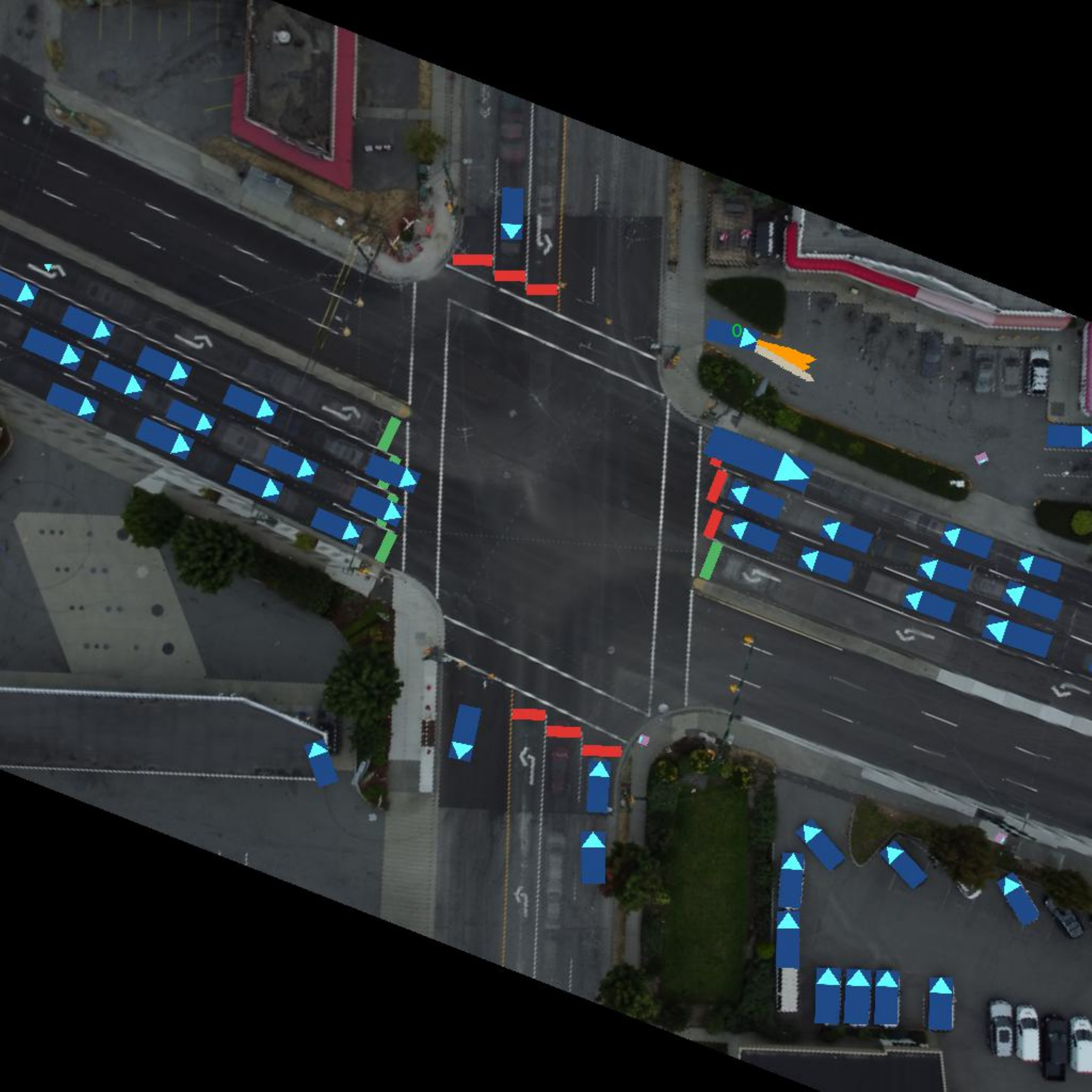}
      \caption{}
    \end{subfigure}
        \begin{subfigure}[b]{0.18\textwidth}
        \centering
        \includegraphics[width=1\linewidth,keepaspectratio]{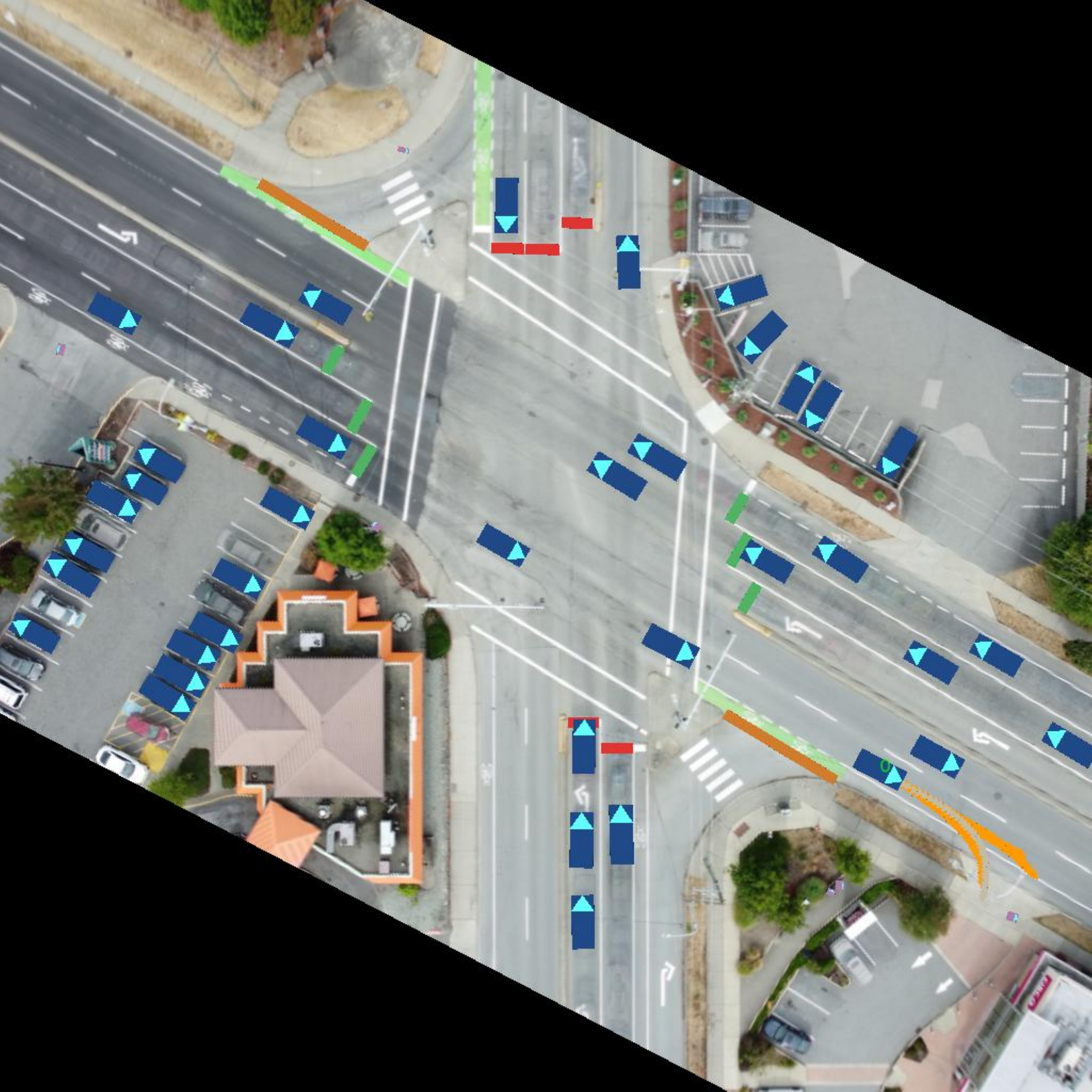}
        \caption{}
    \end{subfigure}
        \begin{subfigure}[b]{0.18\textwidth}
        \centering
        \includegraphics[width=1\linewidth,keepaspectratio]{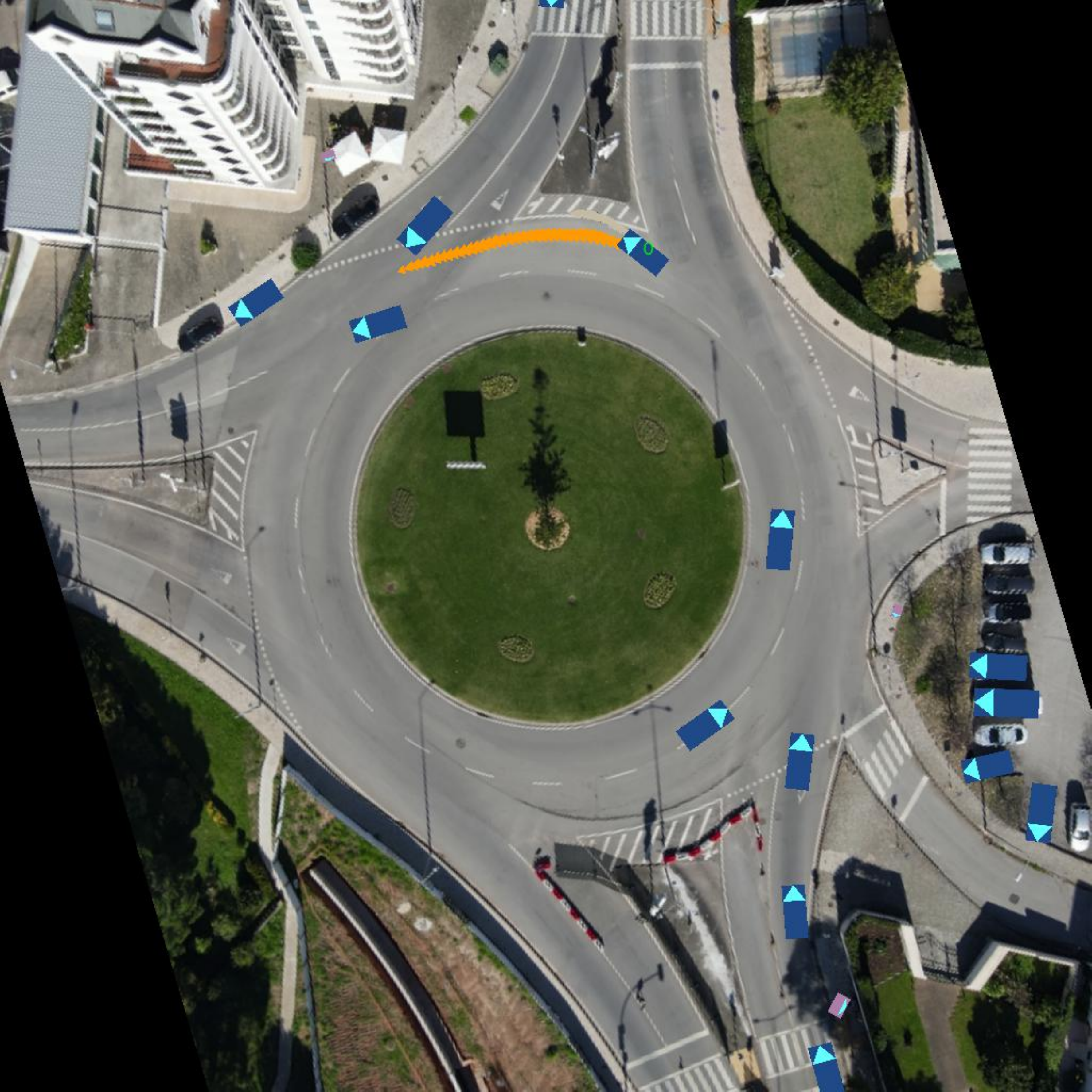}
                \caption{}
    \end{subfigure}
        \begin{subfigure}[b]{0.18\textwidth}
        \centering
        \includegraphics[width=1\linewidth,keepaspectratio]{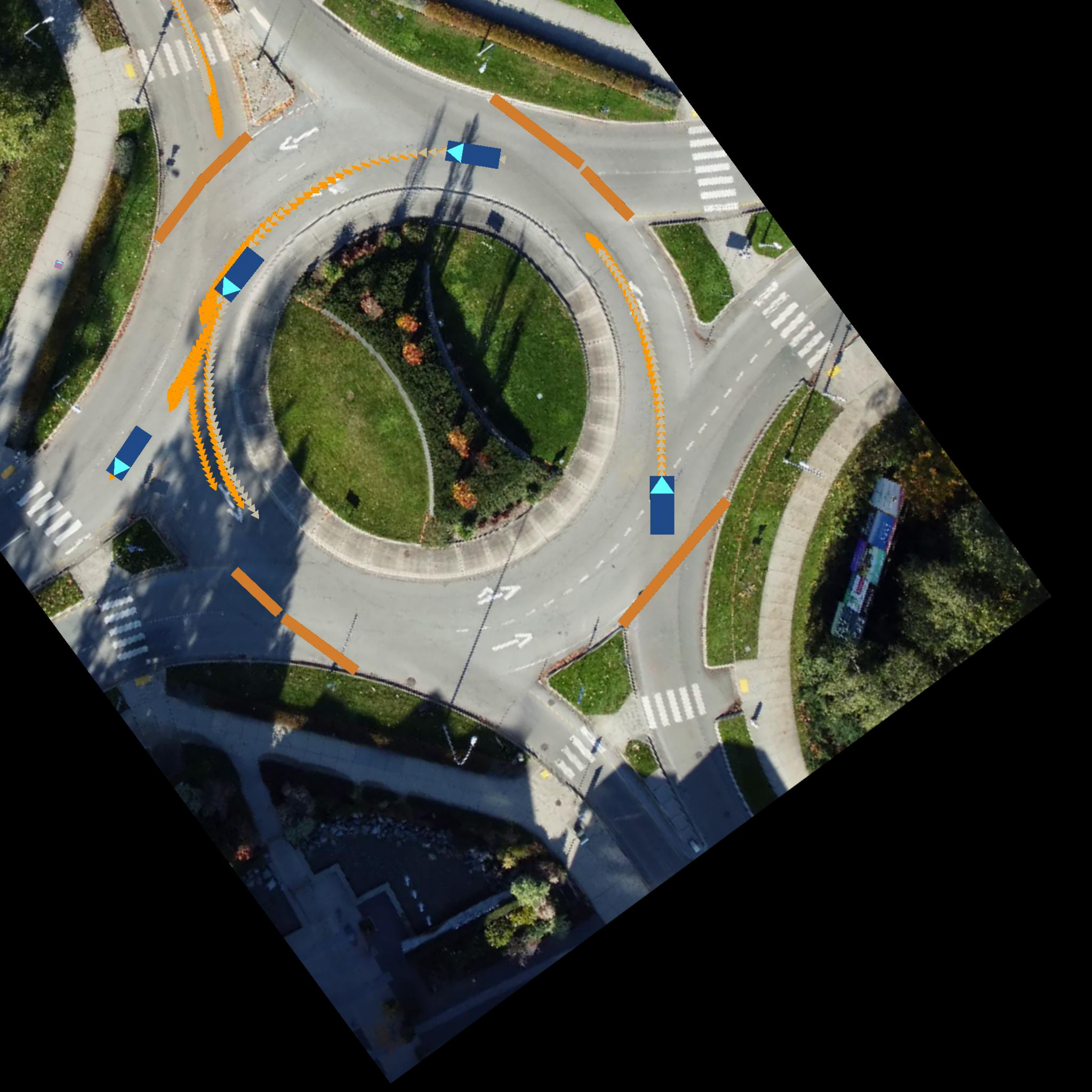}
                \caption{}
    \end{subfigure}
                \begin{subfigure}[b]{0.18\textwidth}
        \centering
        \includegraphics[width=1\linewidth,keepaspectratio]{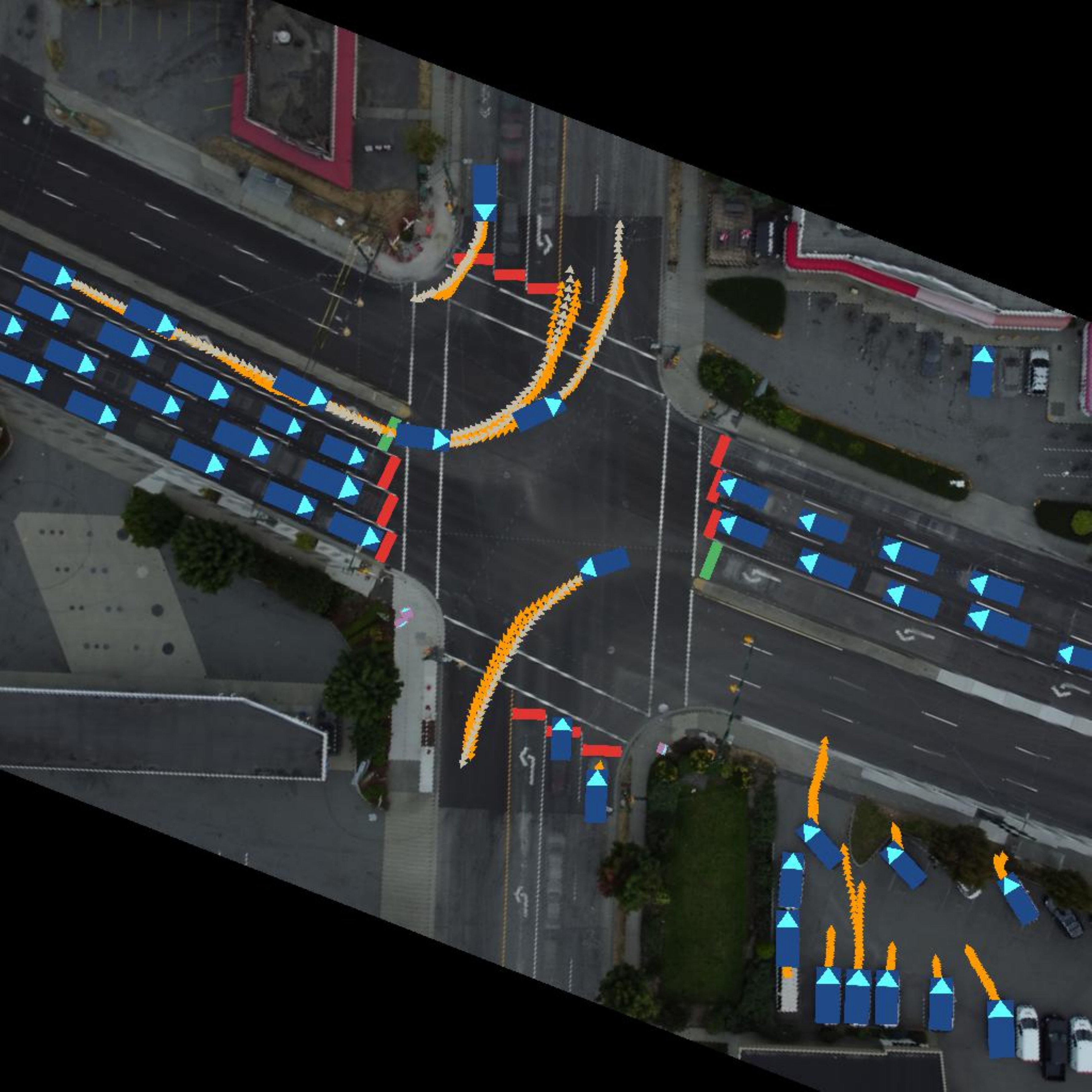}
        \caption{}
    \end{subfigure}
    \caption{\textbf{Column (a-c)} Ego-only trajectory prediction of 40 timesteps based on the observation of only one initial timestep for vehicle agents. We show 10 sampled trajectories in orange alongside the ground truth trajectory colored in grey for both ITRA-V-HDM and ITRA-V-AIM. Note that ITRA trained with the AIM representation generates more realistic samples by leveraging the road context, particularly in scenarios such as entering parking lots. \textbf{Column (d-e)} Examples of Multi-agent trajectory predictions on two different map representations.}
    \label{fig:traj-v-viz}
\end{figure*}

\begin{figure*}[t]
     \centering
     \begin{subfigure}[b]{0.18\textwidth}
        \centering
        \includegraphics[width=1\linewidth,keepaspectratio]{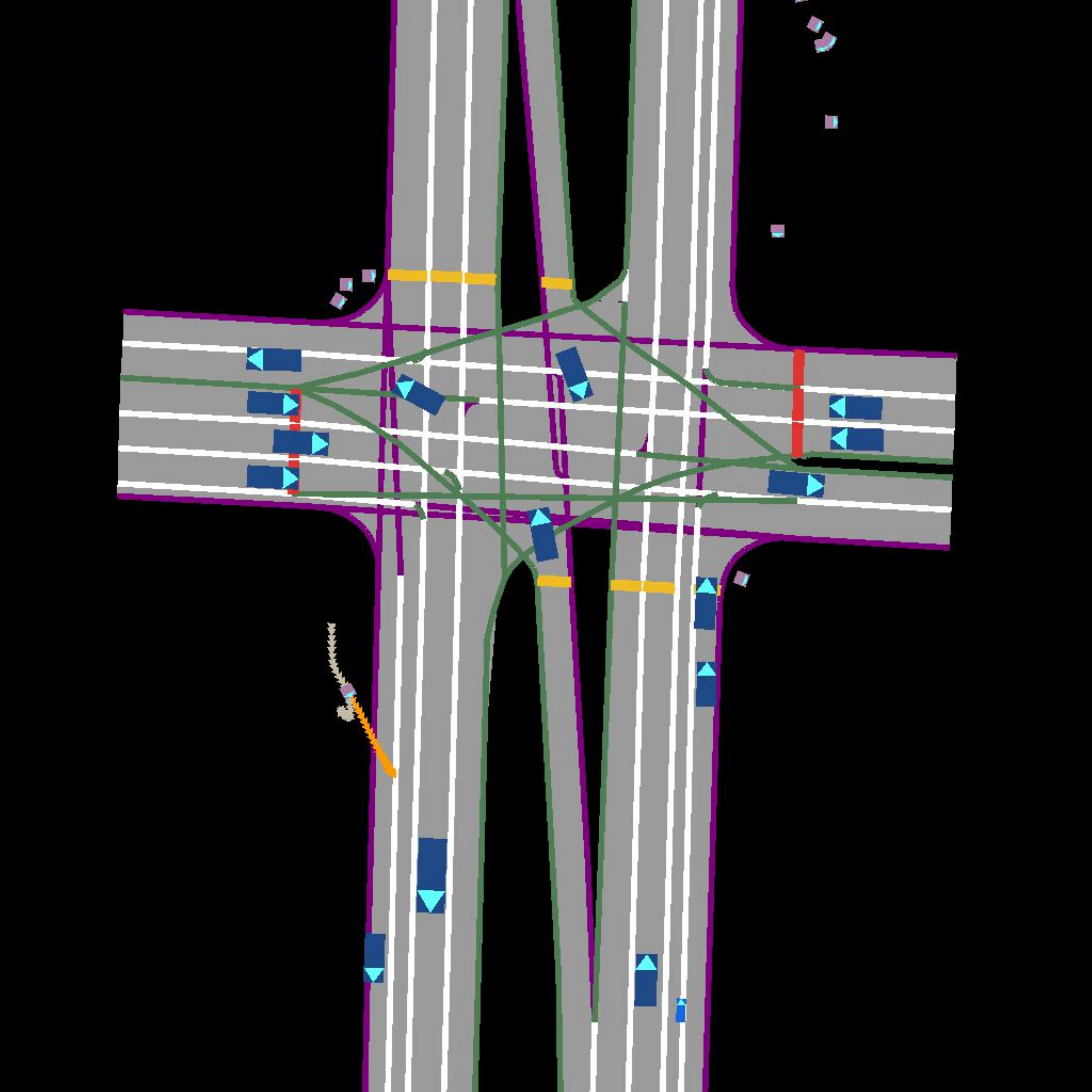}
    \end{subfigure}
    \begin{subfigure}[b]{0.18\textwidth}
        \centering
        \includegraphics[width=1\linewidth,keepaspectratio]{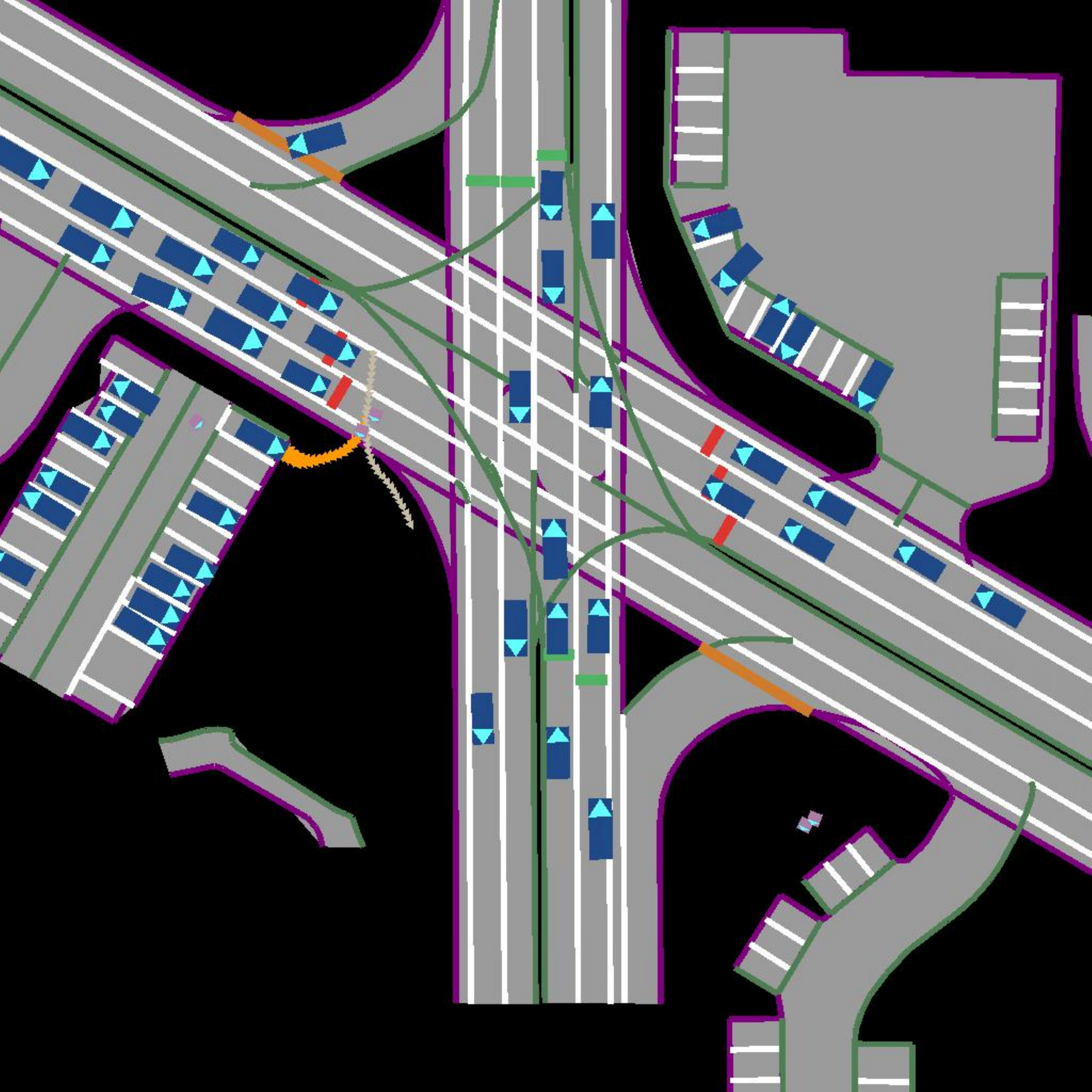}
    \end{subfigure}
    \begin{subfigure}[b]{0.18\textwidth}
        \centering
        \includegraphics[width=1\linewidth,keepaspectratio]{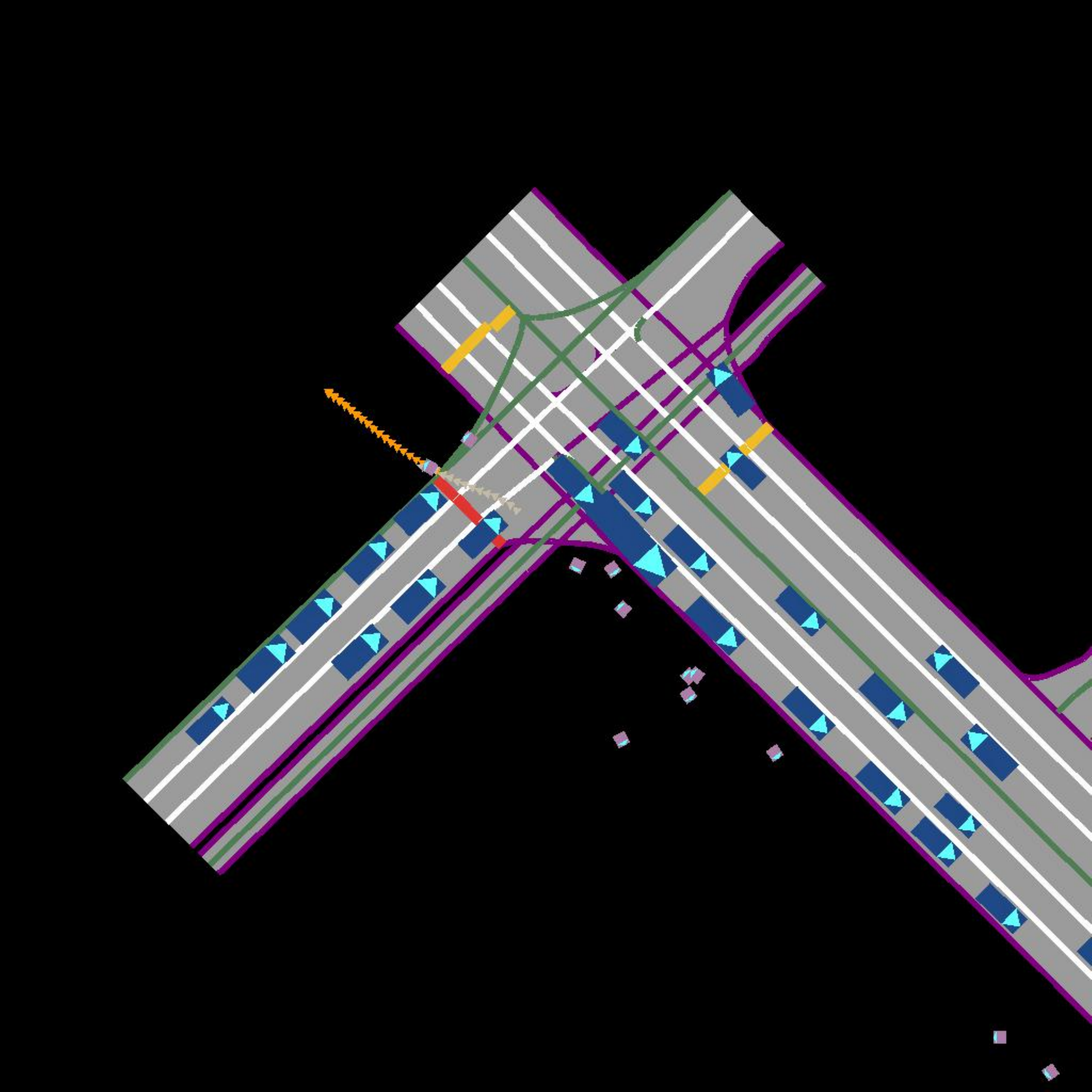}
    \end{subfigure}
        \begin{subfigure}[b]{0.18\textwidth}
        \centering
        \includegraphics[width=1\linewidth,keepaspectratio]{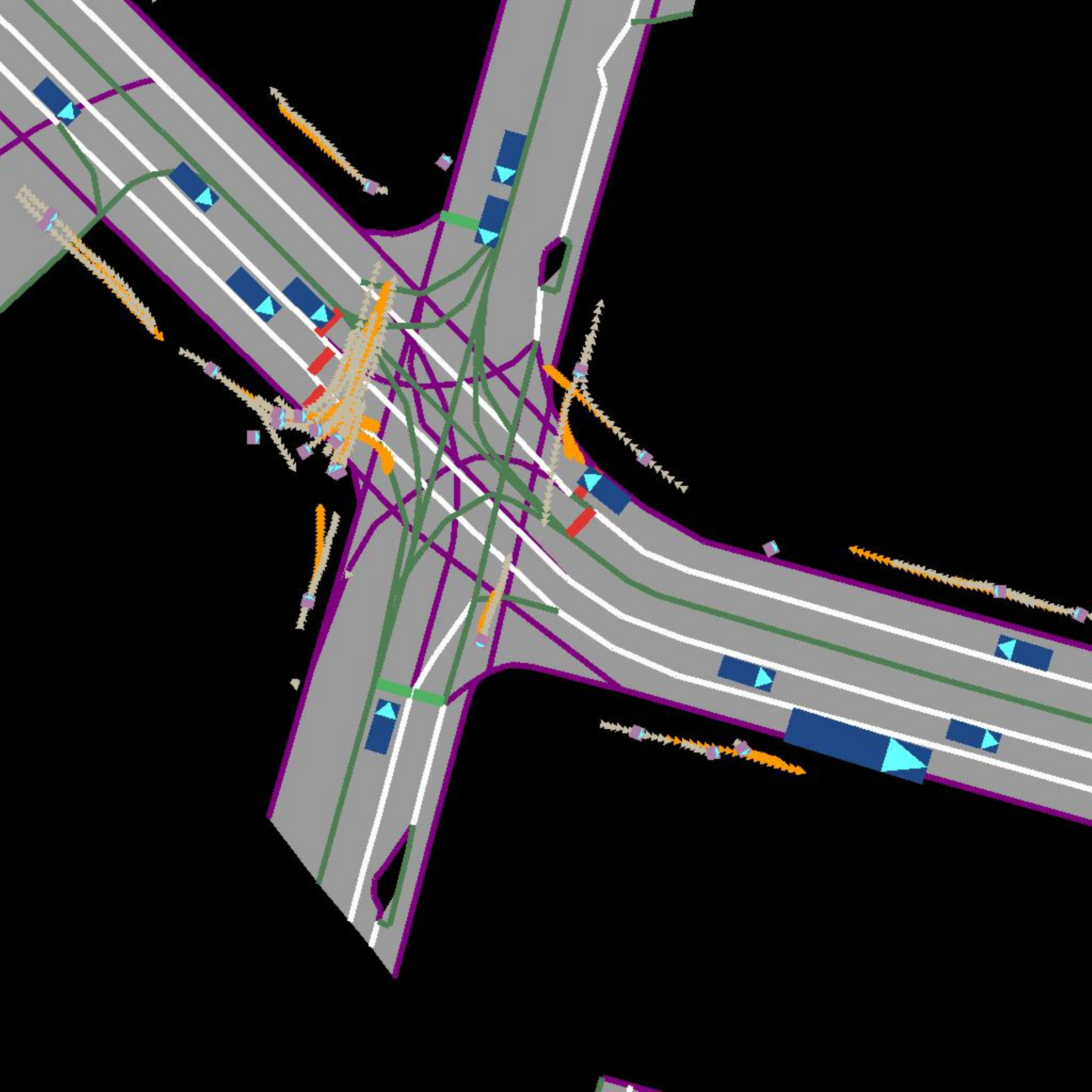}
    \end{subfigure}
        \begin{subfigure}[b]{0.18\textwidth}
        \centering
        \includegraphics[width=1\linewidth,keepaspectratio]{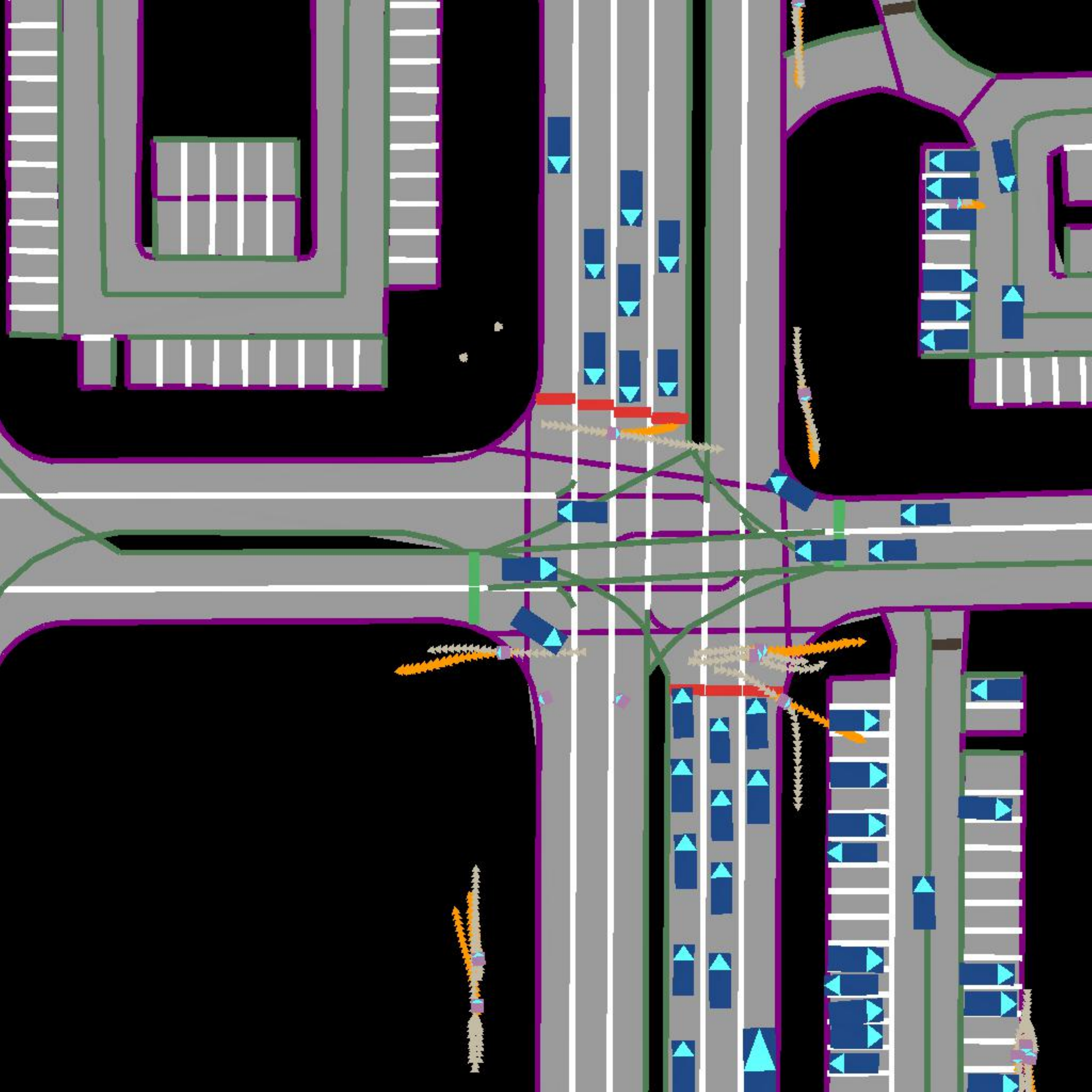}
    \end{subfigure}
            \begin{subfigure}[b]{0.18\textwidth}
        \centering
        \includegraphics[width=1\linewidth,keepaspectratio]{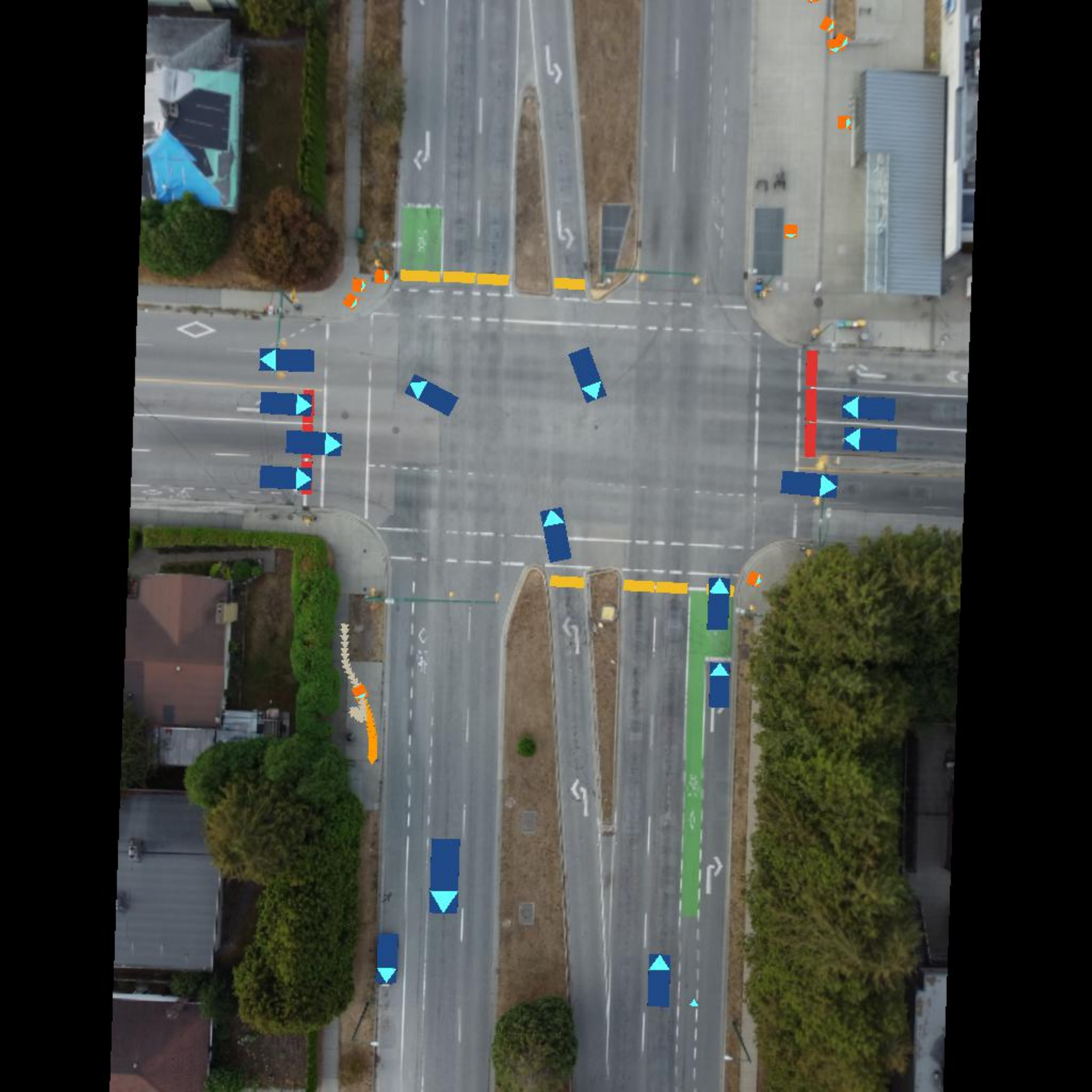}
      \caption{}
    \end{subfigure}
        \begin{subfigure}[b]{0.18\textwidth}
        \centering
        \includegraphics[width=1\linewidth,keepaspectratio]{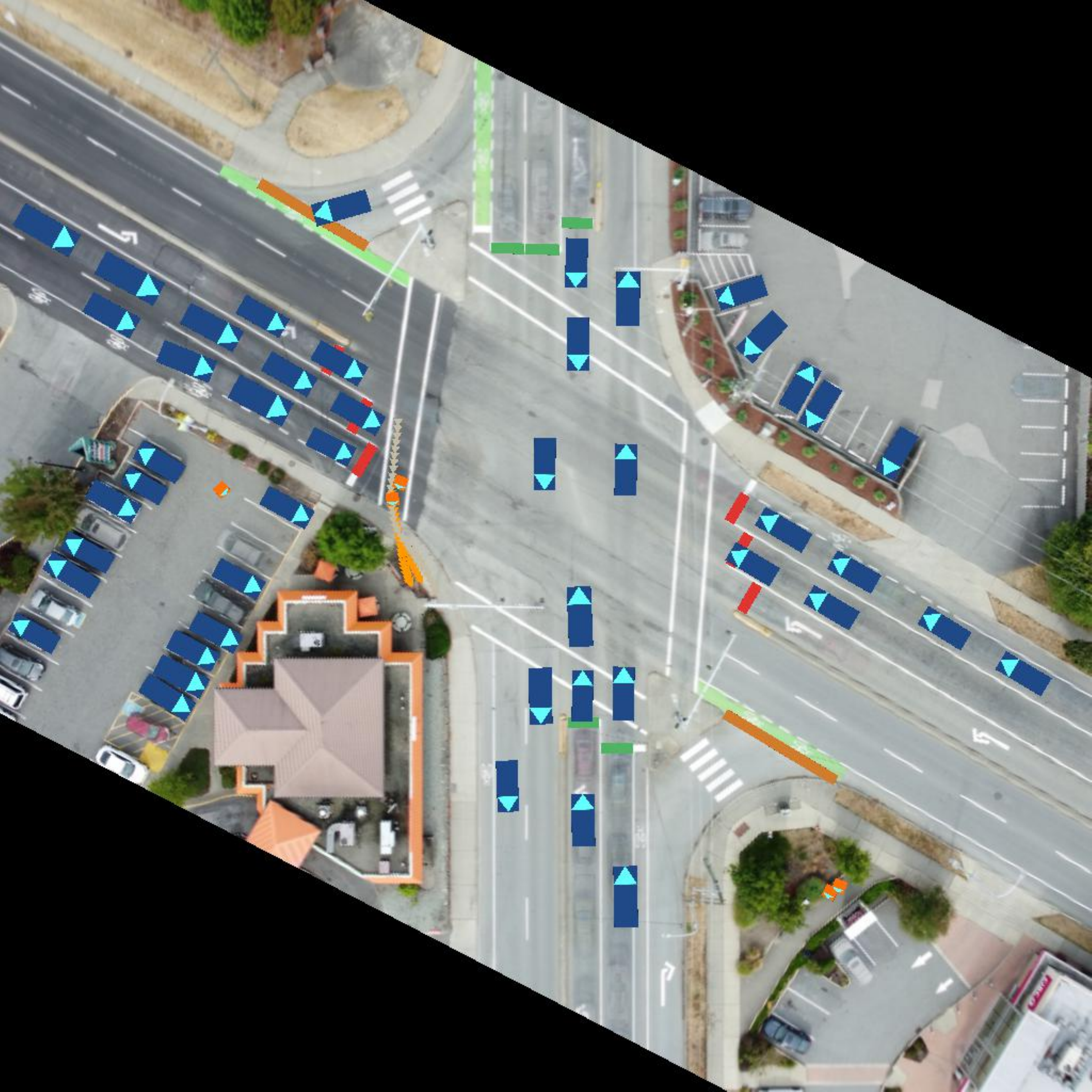}
        \caption{}
    \end{subfigure}
        \begin{subfigure}[b]{0.18\textwidth}
        \centering
        \includegraphics[width=1\linewidth,keepaspectratio]{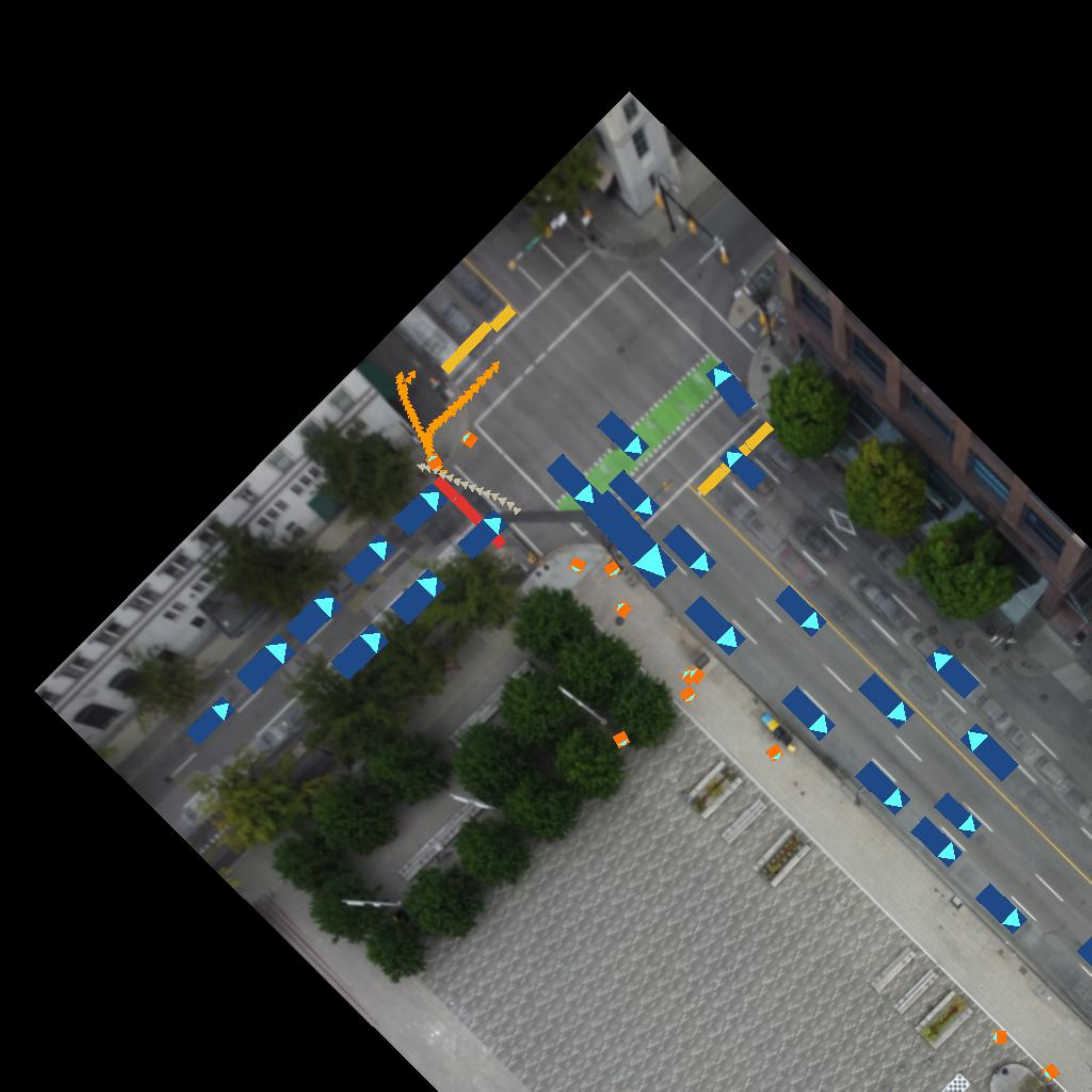}
                \caption{}
    \end{subfigure}
        \begin{subfigure}[b]{0.18\textwidth}
        \centering
        \includegraphics[width=1\linewidth,keepaspectratio]{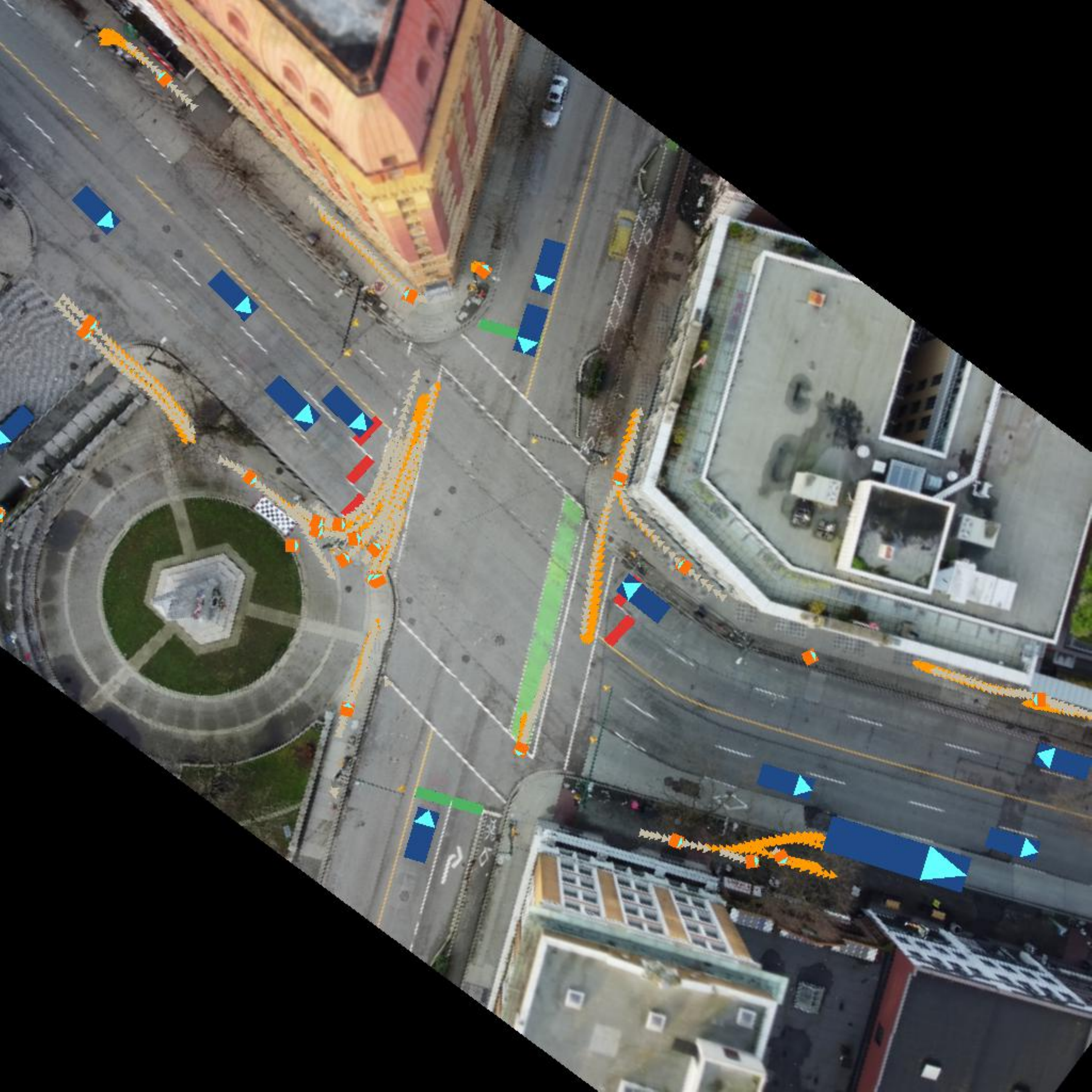}
                \caption{}
    \end{subfigure}
                \begin{subfigure}[b]{0.18\textwidth}
        \centering
        \includegraphics[width=1\linewidth,keepaspectratio]{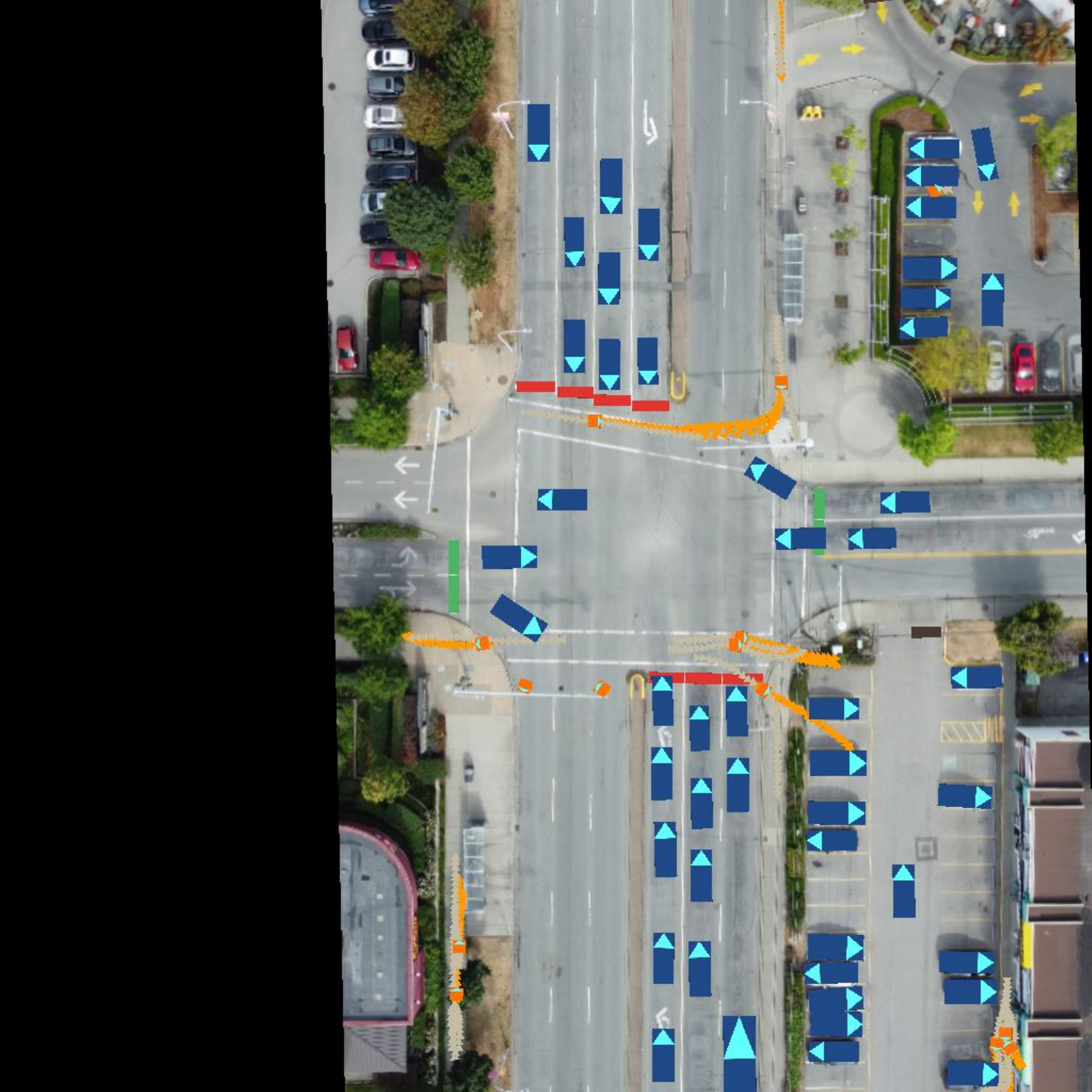}
        \caption{}
    \end{subfigure}
    \caption{\textbf{Column (a-c)} Ego-only trajectory prediction of twelve seconds based on the observation of only one initial observation for pedestrians. We show 5 sampled trajectories in orange alongside the ground truth trajectory colored in grey for both ITRA-P-HDM and ITRA-P-AIM. Pedestrians are highlighted with orange bounding boxes in ITRA-P-AIM for visualization purposes. \textbf{Column (d-e)} Examples of Multi-agent trajectory predictions on two different map representations. Note that the pedestrian model trained with AIM tends to navigate within designated areas such as sidewalks and crosswalks. This behavior is facilitated by the incorporation of comprehensive road context information provided by the AIM, enabling the model to leverage the contextual cues to figure out pedestrians movement patterns.}
    \label{fig:traj-p-viz}
\end{figure*}

We evaluate the effectiveness of our AIM representation with ITRA (ITRA-AIM) on a dataset comprising 5.5 hours of traffic data collected using a commercial drone in 20 locations primarily in Canada. These locations include roundabouts, signalized and unsignalized intersections, and highways, providing a diverse set of road geometries for training and evaluation.  We carefully selected 11 locations out of the 20 locations for training the ITRA-AIM model on vehicles. These particular locations were chosen based on the presence of rich driving behavior. Similarly, we selected 17 locations specifically for training our model on pedestrian data, considering that pedestrians are typically absent from locations such as highways. 

Our dataset comprises 300k four-second long segments sampled at 10 Hz of vehicle data.  On the other hand, the pedestrian dataset consists of approximately 200k segments, each spanning twelve seconds and sampled at a rate of 2.5 Hz. This lower sampling rate is employed due to the comparatively slower movement of pedestrians compared to vehicles, which is consistent with other well-known pedestrian datasets~\cite{lerner2007crowds,pellegrini2009you}. We reserved the final 5\% of our drone recordings obtained at each location as our validation dataset, ensuring that the training data has no causal relationship to the validation data. 

In our ablation studies, we demonstrate the importance of rendering traffic lights in the AIM, as well as the impact of aerial image quality on displacement errors and infraction rates of the generated samples. These findings provide insights to why prior work~\cite{zhang2022trajectory} obtained inferior results on satellite images as mentioned in \cref{sec:related}. We also test on an aerial image from Bing~\cite{bing-maps} at one of our recording locations.
\subsection{Implementation details}
To demonstrate the impact of AIM on motion prediction for pedestrian and vehicle agents, we train separate models for the two agent types (ITRA-V and ITRA-P). In addition, to compare ITRA-AIM with the original ITRA that utilizes the rasterized HD map (ITRA-HDM),  we apply the same training procedure on ITRA-AIM as our baseline, training each component of the network from scratch and using the same training hyper-parameters for ITRA-AIM and ITRA-HDM. We use an identical CNN encoder for encoding AIM which consists of a 4-layer CNN model for our ITRA-AIM model but also experiment with a ResNet-18 backbone on the vehicle dataset to encode the AIM representation, because it contains more information than the rasterized HD map. While the training setup is the same for ITRA-V and ITRA-P, we apply a unicycle dynamics model~\cite{uni} as $f_{kin}$ for pedestrians instead of the bicycle model we use for vehicles.


ITRA adopts classmates forcing~\cite{Tang2019MultipleFP} in the training phase, which provides ground truth states to the model for all agents beside the ego agent. At test time, ITRA trained on the vehicle dataset (ITRA-V) jointly predicts the future trajectories for all vehicles while other agent types like pedestrians are replayed in the scene with ground truth trajectories. Similarly, ITRA-P predicts the motion of all pedestrians in the scene given the ground truth vehicle trajectories. We train all of our models with a random observation length between 1 to 10 timesteps to prevent overfitting on past observations and they are trained until the validation loss converges.  The training time of ITRA-AIM is comparable to that of ITRA-HDM on 4 NVIDIA GeForce RTX 2080 Ti GPUs.

\subsection{Evaluation Metrics}
Common metrics for evaluating trajectory prediction models are ADE and FDE, which measure how close the sampled trajectory is to the ground truth trajectory.  In the multi-agent setting, ADE and FDE are averaged across all $N$ agents. Given $K$ trajectory prediction samples, the generated trajectory with the minimum error is selected for calculating $\text{minADE}_K$ and $\text{minFDE}_K$.  As AIM does not explicitly label the drivable area, we evaluate the prediction performance of the ITRA-V-AIM model on how often it commits off-road infractions. To accomplish this, we calculate the off-road rate using the method described in~\cite{lioutas2021}, assuming access to a drivable surface mesh for evaluation purposes. The computed off-road rate is zero when all four corners of the vehicle are within the drivable area. Furthermore, we also measure the collision rate in a multi-agent prediction setting using the intersection over union (IOU)-based collision metric from~\cite{lioutas2021}. Both the off-road rate and collision rate reported in \cref{tab:v_valid,tab:abl_traffic_light,tab:abl_q_aim} are averaged across the number of agents, samples, and time. Since AIM provides additional road context information like sidewalks and crosswalks, we also measure the diversity of trajectories generated for pedestrians by measuring the Maximum Final Distance (MFD) averaged over number of agents introduced in \cite{9565113}. 

\subsection{Experimental results}
We present our results on the validation dataset for ITRA-P in \cref{tab:p_valid} and for ITRA-V in \cref{tab:v_valid}. In the case of ITRA-P, we jointly predict the eight-second future given the initial four-second observation. As for ITRA-V, we predict the trajectory for a time horizon of 40 timesteps (four seconds) while observing the first 10 timesteps. On the pedestrian dataset, ITRA-P-AIM outperforms ITRA-P-HDM across all evaluation metrics and achieves higher diversity. This result indicates the importance of providing scene context information for pedestrians when modeling them in traffic simulations. We showcase our predicted examples on the validation dataset in \cref{fig:traj-p-viz}. ITRA-V-AIM demonstrates competitive performance in reconstructing ground truth trajectories on the validation dataset. Regarding the off-road rate, ITRA-V-AIM with a ResNet-18 backbone matches the performance of ITRA-V-HDM while maintaining similar displacement errors. \Cref{fig:traj-v-viz} presents validation examples comparing ITRA-V-HDM and ITRA-V-AIM on the vehicle dataset.

\subsection{Ablation Studies}
\label{sec:ablation}
To investigate the impact of incorporating traffic light states into the AIM representation, we analyze their influence on prediction results, since previous work \cite{zhang2022trajectory} did not evaluate representations that include traffic light states. Specifically, we select locations from our dataset where traffic light states are available and conduct a comparative evaluation of the ITRA-V-AIM model on AIMs with and without traffic light states.

Our results, presented in Table \ref{tab:abl_traffic_light}, demonstrate a notable reduction in collision rates of over 15\% and a 12\% decrease in the minADE metric when traffic light states are rendered on AIMs. Additionally, we observe a similar performance improvement for our pedestrian model, ITRA-P-AIM, when traffic light states are incorporated. We attribute part of the AIM's competitive performance to our proposed image-texture-based differentiable rendering module, which enables the integration of traffic light states within the AIM. 
\begin{figure}[t]
     \centering
     \begin{subfigure}[b]{0.45\columnwidth}
         \centering
         \includegraphics[width=\columnwidth]{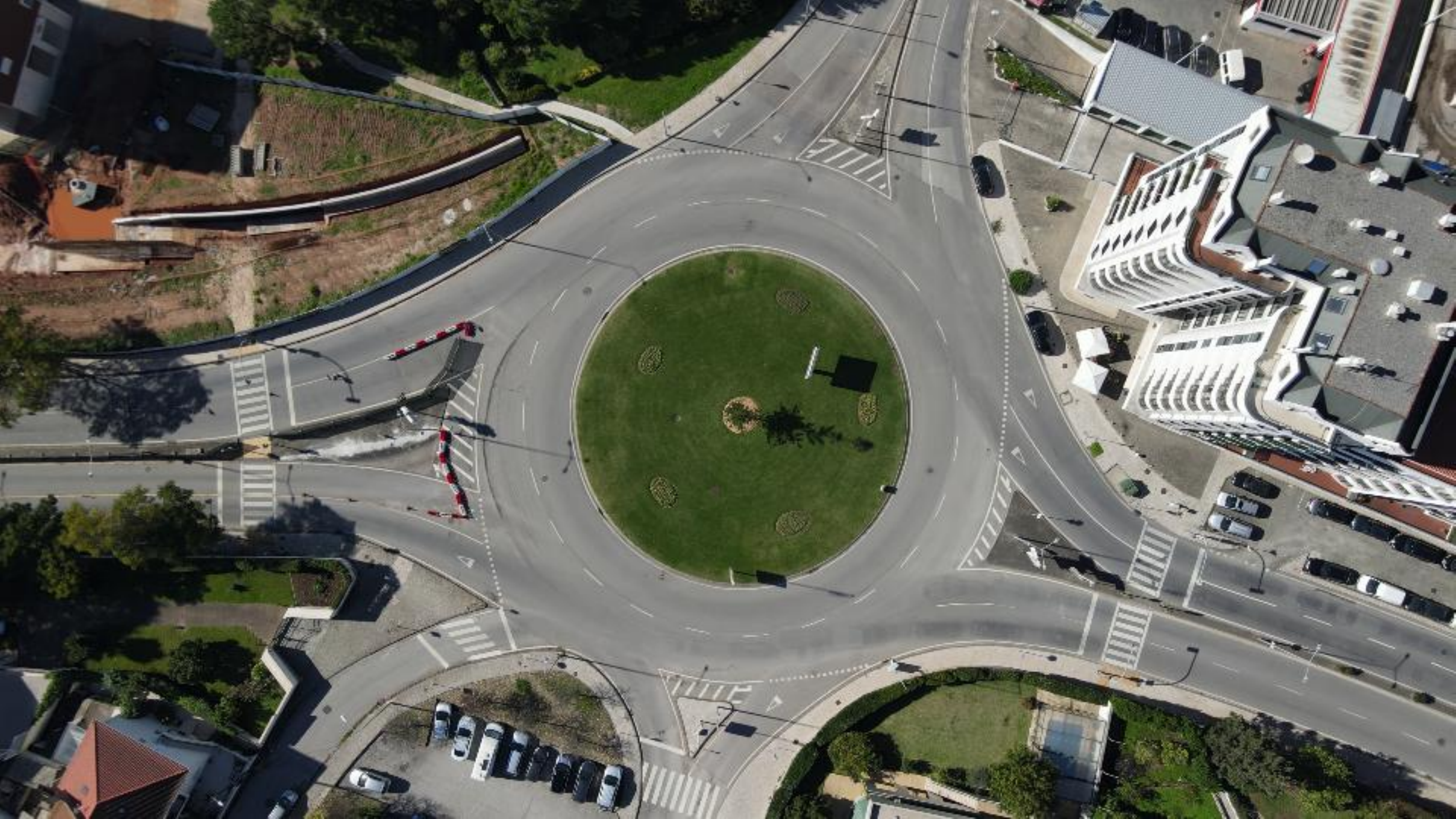}
         \caption{Original aerial image}
         \label{fig:abl_locaitons_org}
     \end{subfigure}
     \begin{subfigure}[b]{0.45\columnwidth}
         \centering
         \includegraphics[width=\columnwidth]{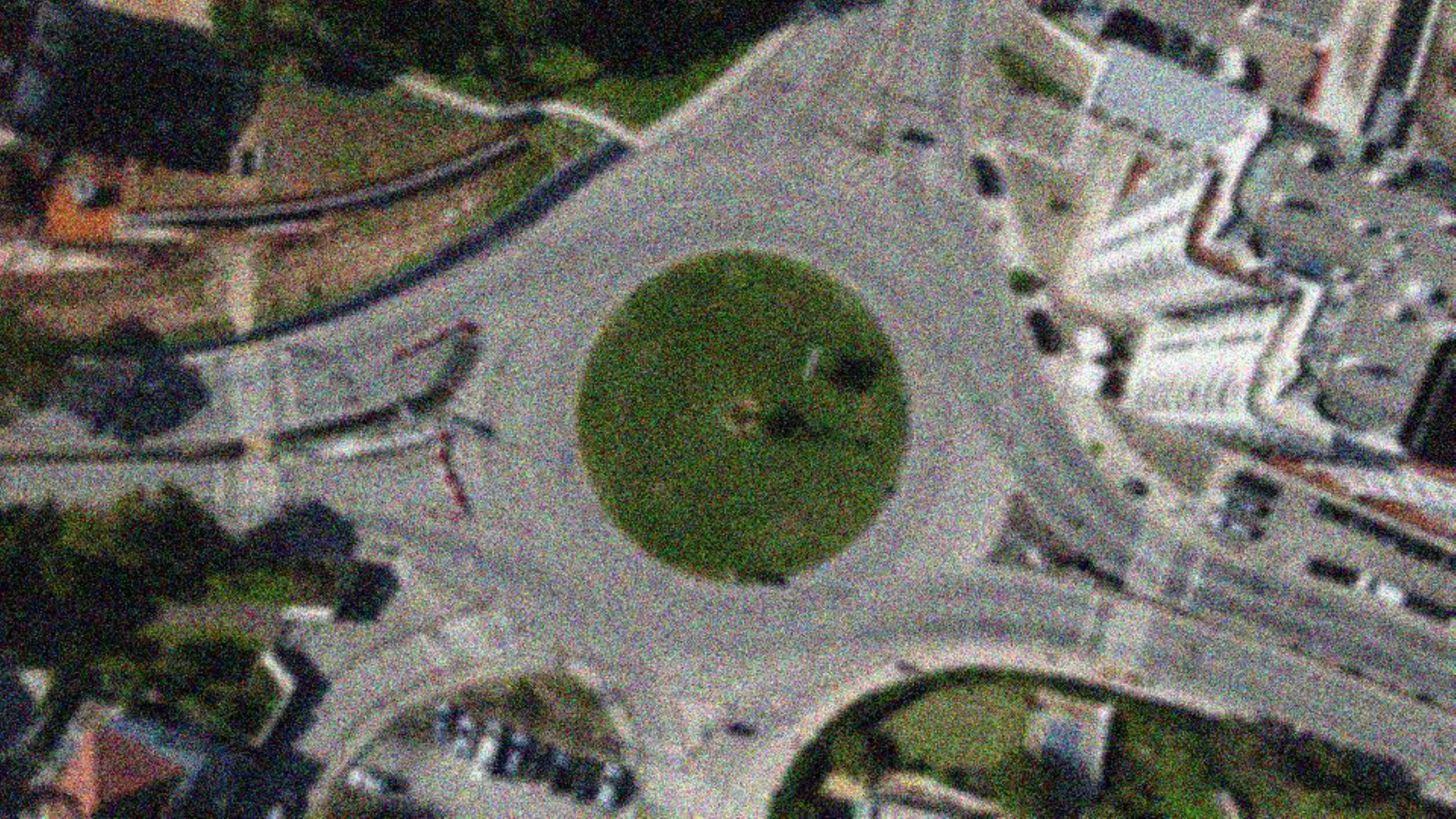}
         \caption{Degraded aerial image}
         \label{fig:abl_locaitons_degra}
     \end{subfigure}
     \caption{Comparison of aerial image quality.}
\end{figure}

We also study the effect of aerial image quality on prediction results. Depending on 
the height at which the image is captured and camera's specifications, the image quality, in terms of resolution and noise level, can vary drastically. To simulate a deterioration in these quality factors, we first apply Gaussian blur to our original aerial images and then add Gaussian noise with a standard deviation of 0.2 to these blurred images (an example is shown in \cref{fig:abl_locaitons_degra}). We evaluate ITRA-V-AIM on the degraded AIMs using the validation agent tracks and report the results in \cref{tab:abl_q_aim}. The degraded AIMs result in increased prediction errors and nearly double the off-road infraction rate. Although the simulated degradation may not precisely emulate the various aspects of the reduction in image quality, our findings highlight the crucial role of capturing high fidelity imagery in ensuring the
accuracy of multi-agent trajectory prediction through our AIM representation.

\begin{table}[t]
    \centering
    \caption{Comparison between with and without traffic light rendered on locations have traffic light labels.}
    \resizebox{\columnwidth}{!}{%
    \begin{tabular}{l||cc}
    \multicolumn{1}{c}{}&\multicolumn{2}{c}{ITRA-V-AIM}\\
        \hline
        Metrics & With traffic lights & Without traffic lights \\
        \hline
        $\text{minADE}_6$$\downarrow$ &\bf{0.37} & 0.42\\
        $\text{minFDE}_6$$\downarrow$ & \bf{0.77} &0.88   \\
        Off-road rate$\downarrow$ &\bf{0.006} &  0.008\\
                Collision rate$\downarrow$ & \bf{0.011}&0.013  \\
        \hline
    \end{tabular}%
    }
    \label{tab:abl_traffic_light}
\end{table}

\begin{table}[t]
    \centering
    \caption{The impact of aerial image quality on prediction results.}
    \resizebox{\columnwidth}{!}{%
    \begin{tabular}{l||cc}
     \multicolumn{1}{c}{}&\multicolumn{2}{c}{ITRA-V-AIM}\\
     \hline
     Metrics & Original AIM & Blurred and noise-added AIM\\
     $\text{minADE}_6$& \bf{0.45}& 0.49\\
     $\text{minFDE}_6$ & \bf{0.93}&1.04\\
     Off-road rate$\downarrow$& \bf{0.008}& 0.015\\ 
     Collision rate$\downarrow$& \bf{0.012}& 0.015\\
        \hline
    \end{tabular}%
    }
    \label{tab:abl_q_aim}
\end{table}

\begin{figure}[t]
     \centering
    \resizebox{0.9\columnwidth}{!}{%
     \begin{subfigure}[b]{0.45\columnwidth}
         \centering
         \includegraphics[width=\columnwidth]{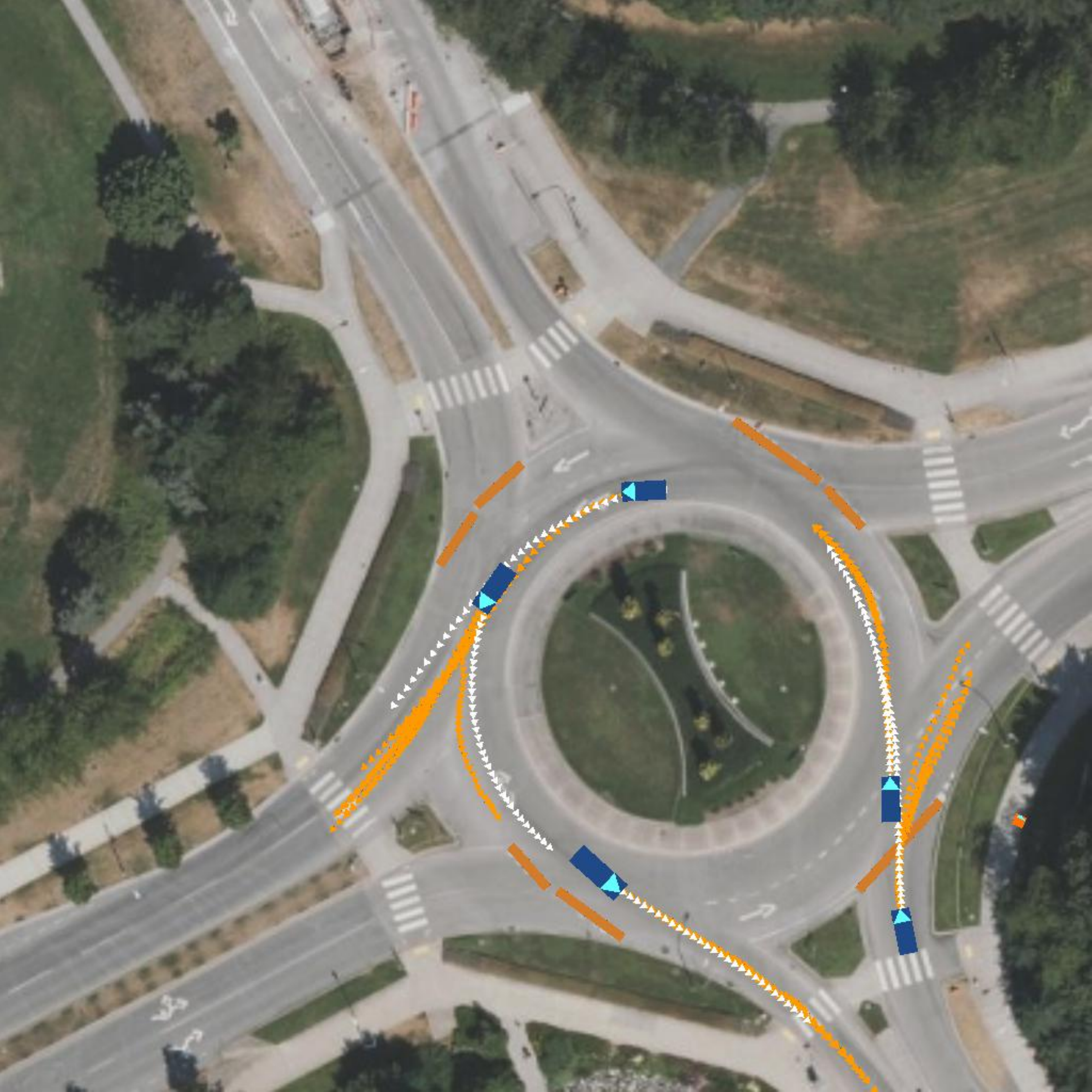}
     \end{subfigure}
     \begin{subfigure}[b]{0.45\columnwidth}
         \centering
         \includegraphics[width=\columnwidth]{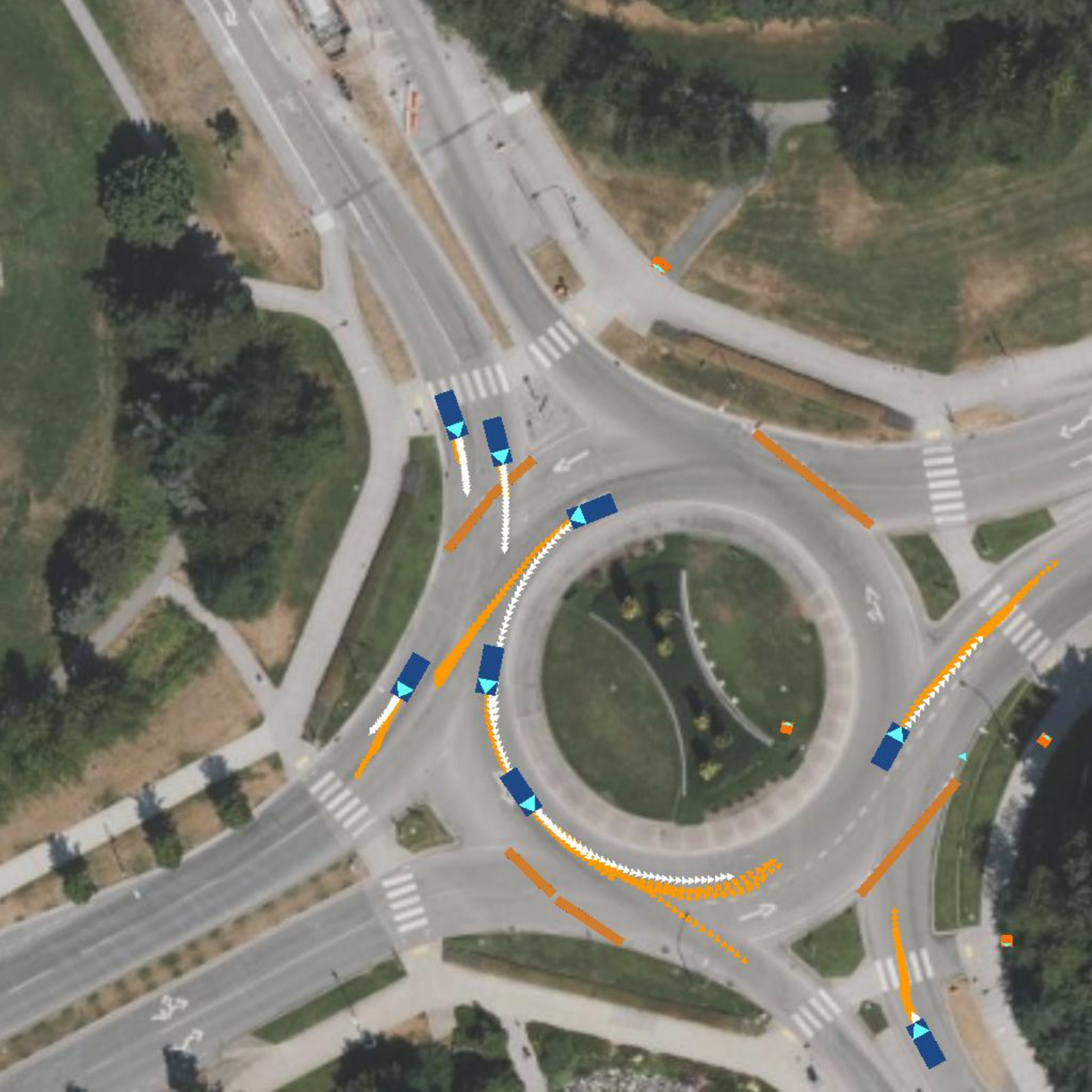}
     \end{subfigure}%
     }
     \caption{Evaluation of the ITRA-V-AIM on a Bing aerial image. The white trajectories represent the ground truth trajectories obtained from validation segments at the same location, and the orange trajectories display 5 sampled trajectories of the ITRA-V-AIM model. Our ITRA-V-AIM model predicts road context-aware trajectories on this aerial image which has different lighting and shading conditions compared to the training AIM. }
     \label{fig:abla}
\end{figure}
\vspace{-1pt}

To demonstrate the flexibility of our AIM representation, we acquired an aerial image from Bing aerial imagery of the same location as \cref{fig:traj-v-viz}d to construct a larger AIM. Despite having different image conditions (such as shading and lighting) compared to our original AIM, our trained ITRA-V-AIM achieves good prediction performance on this larger map with validation data segments as shown in \cref{fig:abla}.

\vspace{-1pt}
\section{Conclusion}
In this work, we have addressed a critical bottleneck in scaling the dataset size for behavioral models used for simulating realistic driving. Specifically, the manual annotation of high-definition maps on new locations impedes progress towards the fully automated labeling of datasets. Rather than pursuing the development of automated map labeling tools, which may introduce additional labeling noise, our proposed aerial image-based map (AIM) requires minimal annotations.  By employing the AIM in a driving simulator through our image-texture-based differentiable rendering, we illustrate that the AIM provides rich road context information for multi-agent trajectory prediction which resulted in more realistic samples for both vehicles and pedestrians. Our results on pedestrian trajectory prediction indicates a substantial improvement when utilizing our AIM representation compared to the baseline representation. In addition, our image-texture-based differentiable rendering module can be easily integrated into any existing behavioral prediction models that consume bird's-eye view images as part of the agents' state representation. While our work has yielded promising results, there are still opportunities remaining for further improvement. These opportunities include an in-depth investigation of semantic or structure-based encoders for the AIM representation to improve on the off-road metrics. Finally, while we constructed AIMs from drone recordings, they could also be employed when collecting data from the vehicle by utilizing existing aerial imagery, although additional effort would be required to align such images to the extracted agent tracks.

\section{ACKNOWLEDGEMENT}
We acknowledge the support of the Natural Sciences and Engineering Research Council of Canada (NSERC), the Canada CIFAR AI Chairs Program, and the Mitacs Accelerate program in partnership with Inverted AI. Additional support was provided by UBC's Composites Research Network (CRN), and Data Science Institute (DSI). This research was enabled in part by technical support and computational resources provided by WestGrid (www.westgrid.ca), Compute Canada (www.computecanada.ca), and Advanced Research Computing at the University of British Columbia (arc.ubc.ca).

\bibliographystyle{IEEEtran}
\bibliography{egbib}

\end{document}